\documentclass[11pt,a4paper]{article}
\usepackage{color}
\usepackage[colorlinks=true]{hyperref}
\usepackage[margin=1in]{geometry}
\usepackage{fullpage,mathrsfs,graphicx,framed,color,amssymb,amsmath,amsthm}
\usepackage[boxed, noline, ruled, linesnumbered]{algorithm2e}
\usepackage{epsfig, graphicx}
\usepackage{latexsym,amsfonts,amsbsy,amssymb}
\usepackage{amsmath,amsthm}
\usepackage{enumerate}
\usepackage{cleveref}
\usepackage{tikz}
\usepackage{stmaryrd} 
\usepackage{pgfplots}
\pgfplotsset{compat=1.17}
\usepackage{multirow}
\usepackage{booktabs}
\usepackage{arydshln} 
\usepackage{bbding} 

\usepackage{changes} 

\usepackage[utf8]{inputenc} 

\usepackage{hyperref}
\usepackage{caption}
\makeatletter 

\@addtoreset{equation}{section}
\makeatother 
\allowdisplaybreaks 
\newtheorem{theorem}{Theorem}[section]

\newtheorem{remark}{Remark}[section]

\allowdisplaybreaks 

\numberwithin{equation}{section}

\begin{document}
	
\title{Solving Time-Fractional Partial Integro-Differential Equations Using Tensor Neural Network}
\author{Zhongshuo Lin\footnote{School of Mathematical Science, Peking University, Beijing 100871, China, and Peking University Chongqing Research Institute of Big Data, Chongqing 401329, China (linzhongshuo@pku.edu.cn).},\ \ \ 
Qingkui Ma\footnote{School of Mathematics and Statistics \& 
Hubei Key Laboratory of Mathematical Sciences,
Central China Normal University, Wuhan 430079, China.,
\textsuperscript{*}Corresponding author (maqingkui@mails.ccnu.edu.cn).}\textsuperscript{*},\ \ \ 
Hehu Xie\footnote{SKLMS, NCMIS, Academy of Mathematics and Systems Science,
Chinese Academy of Sciences, No.55, Zhongguancun Donglu, Beijing 100190, 
China, and School of Mathematical Sciences, University of Chinese Academy
of Sciences, Beijing, 100049, China (hhxie@lsec.cc.ac.cn).}\ \ \ and \ \ 
Xiaobo Yin\footnote{School of Mathematics and Statistics \& Key Laboratory of Nonlinear Analysis \& Applications (Ministry of Education),
Central China Normal University, Wuhan 430079, China (yinxb@mail.ccnu.edu.cn).\\ 
 The work was supported by the Strategic Priority Research Program of the Chinese Academy of Sciences (XDA0480504, XDB0620203, XDB0640300, XDB0640000), National Natural Science Foundation of China (Nos. 12471380, 1233000214), Science Challenge Project (TZ2024009), National Key Laboratory of Computational Physics (No. 6142A05230501), and National Center for Mathematics and Interdisciplinary Science, CAS.
}
}

\date{}
\maketitle
\begin{abstract}
In this paper, we propose a novel machine learning method based on tensor neural network subspace to solve  linear time-fractional diffusion-wave equations and nonlinear time-fractional partial integro-differential equations. Within this framework, the trial function is formulated as the tensor neural network function multiplied by $t^{\mu}$ where the exponent $\mu$ is chosen according to the fractional-order operator. The Gauss-Jacobi quadrature is then employed to accurately evaluate the temporal Caputo derivative. Moreover, an alternating optimization strategy is adopted during the training process inspired by the concept of neural network subspace. Finally, several numerical examples are provided to validate the efficiency and accuracy of the proposed tensor neural network based machine learning method. 


\end{abstract}

\section{Introduction}
Fractional partial differential equations (FPDEs) differ from traditional differential equations in their nonlocal nature induced by fractional differential operators. This nonlocality enables FPDEs to effectively 
capture memory effects and hereditary characteristics which are widely observed 
in real-world phenomena \cite{ma2023pmnn,pang2019fpinns,jiang2011high}. 
Due to these properties, FPDEs demonstrate significant potential 
for applications in diverse fields \cite{sun2020fractional}, 
including anomalous diffusion, fluid mechanics, electromagnetism, 
statistical modeling, and signal processing. For example, the time-fractional diffusion wave equation can simulate complex 
diffusion and wave phenomena in diverse applications such as fluid dynamics, petroleum reservoir engineering, and anomalous transport processes \cite{huang2013two,jiang2017fast,srivastava2010multi,torvik1984}.

However, solving FPDEs faces a huge challenge. Analytical solutions are often 
intractable and difficult to derive, especially for problems involving complex geometries, 
boundary conditions, or nonlinear terms, and even when available, 
frequently involve complex special functions, 
such as Mittag-Leffler functions, H-functions, and Wright functions \cite{admon2023new}.
To address these challenges, various numerical methods have been developed to solve FPDEs, 
e.g., the finite difference method \cite{kumari2023single,she2022transformed,sun2020fractional}, 
finite element method \cite{jiang2011high, zhang2024fast,zhao2017galerkin}, 
spectral method \cite{dehghan2016legendre, harizanov2020survey}, 
wavelet method \cite{jafari2011application,mohammadi2018generalized}, 
Laplace transform method \cite{salahshour2012solving}, 
matrix method \cite{talaei2018operational}, 
and domain decomposition method \cite{momani2006analytical, momani2007numerical,ray2005approximate}. 
These traditional approaches have successfully tackled fractional 
derivatives to some extent, laying a solid foundation for the further development of numerical methods for FPDEs.

While these achievements are encouraging, traditional numerical techniques often 
struggle to efficiently handle the nonlocal nature of fractional derivatives, 
frequently resulting in increased computational complexity, memory consumption 
and reduced accuracy. To fully exploit the potential of FPDEs in various fields, 
it is crucial to develop more efficient and accurate numerical methods. 
In recent years, neural networks have become an active research 
focus across various fields of science and engineering. Thanks to their powerful 
approximation capabilities, neural networks have demonstrated exceptional performance 
in solving complex problems. 
Many neural-network-based numerical methods have been proposed to 
solve partial differential equations. 
For instance, Lagaris et al. \cite{lagaris1998artificial} are early pioneers in applying 
artificial neural networks to solve differential equations, constructing approximate 
solutions for initial and boundary value problems. 
Raissi et al. \cite{raissi2019physics} introduce the physics-informed 
neural networks (PINNs), a novel general-purpose function approximator 
that efficiently encodes physical information into the loss function 
of the neural network for guiding training. The forward and inverse 
problems of partial differential equations were successfully solved 
within a given data set.

To solve space-time fractional advection-diffusion equations, 
fractional physics-informed neural networks (fPINNs) have been proposed \cite{pang2019fpinns}. 
In fPINNs, the $\mathrm{L1}$ scheme is employed to approximate the 
temporal Caputo fractional derivative, while the shifted vector Gr\"{u}nwald-Letnikov 
scheme is used to discretize  the spatial fractional Laplacian. 
Ma et al. introduce Bi-orthogonal fPINN \cite{ma2023bi}, which combines biorthogonal 
constraints with PINNs to solve random FPDEs.
Ma et al. \cite{ma2023pmnn} establish a physical model-driven neural 
network to solve FPDEs, which effectively combines deep neural networks (DNNs) 
with interpolation approximation of fractional derivatives. 
Ren et al. \cite{ren2023class} propose an 
improved fractional physics-informed neural networks (IFPINNs), 
which enhance the capabilities of traditional PINNs by 
incorporating nonlinear weight propagation to solve FPDEs with high oscillations or 
singular perturbations. Zhang et al. \cite{zhang2024adaptive} establish 
an adaptive loss weighting auxiliary output fractional physics-informed neural 
networks (AWAO-fPINNs) to solve time-fractional partial integral-differential equations (TFPIDEs). Guo et al. \cite{guo2022monte} propose Monte Carlo fractional physics-informed neural networks (MC-fPINNs) for solving high-dimensional FPDEs. By implementing an unbiased Monte Carlo sampling strategy to approximate fractional derivatives in neural network outputs, their framework reduces computational costs while maintaining accuracy compared to fPINNs. Hu et al. \cite{hu2024tackling} propose an improved Monte Carlo fractional/tempered fractional PINNs (MC-fPINNs/MC-tfPINNs) framework to overcome the curse of dimensionality in FPDEs. By replacing 1D Monte Carlo sampling with Gauss-Jacobi quadrature for fractional derivative approximation, their method significantly reduces variance and accelerates convergence. It is also worth mentioning  that, Wang and Karniadakis \cite{wang2024gmc} propose a general (quasi) Monte Carlo PINN to solve FPDEs on irregular domains. Zhang et al. \cite{zhang2025spectral} proposed spectral-fPINNs, a framework combining spectral methods with PINNs that employs Jacobi polynomials for global discretization of FPDEs.

Recently, Wang et al. \cite{li2024tensor,liao2022solving,wang2022tensor,wang2024solving,wang2024computing} 
propose a type of tensor neural network (TNN)  
and corresponding machine learning method to solve high-dimensional problems. 
The special structure of TNN transforms the high-dimensional integration of TNN functions involved in the loss functions to one-dimensional integration, 
which can be computed using the classical quadrature schemes with high accuracy and efficiency.  
The TNN-based machine learning method has already been used to solve 
high-dimensional eigenvalue problems and boundary value problems by 
Ritz-type loss functions \cite{wang2022tensor}. Furthermore, in \cite{wang2024computing}, 
the multi-eigenpairs are computed by a machine learning method which combines TNN and Rayleigh-Ritz process. With the help of TNN, a posterior error estimator 
can be used as a loss function in machine learning methods to solve high-dimensional 
boundary value problems and eigenvalue problems \cite{wang2024solving}.
TNN has also been utilized to solve the Schrödinger equation for 20,000-dimensional coupled harmonic oscillators \cite{2023Tackling}, 
high-dimensional Fokker-Planck equations \cite{wang2025tensor}, and high-dimensional time-dependent problems\cite{kao2024petnns}.

In this paper, we aim to design efficient machine learning method for solving the following (linear or nonlinear) TFPIDEs with single-term or multi-term time fractional derivative: Find $u(x,t)$ such that 
\begin{equation} \label{FPIDE_problem}
	\left\{ \begin{aligned}
		Lu(\boldsymbol{x},t)&=f(\boldsymbol{x},t,u(\boldsymbol{x},t)),  
		(\boldsymbol{x},t)\in \Omega \times (0,T],\\
		u(\boldsymbol{x},t)&=g(\boldsymbol{x},t),  (\boldsymbol{x},t)\in \partial \Omega \times (0,T],\\
		u(\boldsymbol{x},0)&=s(\boldsymbol{x}),  \boldsymbol{x}\in \Omega,
	\end{aligned} \right. 
\end{equation}
where the domain $\Omega$ is defined as  $(a_1, b_1)\times\cdots \times (a_d,b_d)$.  
The function $f(\boldsymbol{x},t,u(\boldsymbol{x},t))$ is linearly or nonlinearly dependent on $u(\boldsymbol{x},t)$.
For the nonlinear case, we consider the form $f(\boldsymbol{x},t,u)=h(\boldsymbol{x},t) -u^2(\boldsymbol{x},t)$ with $h(\boldsymbol{x},t)$ 
given. Here $L$ represents an operator defined as follows:
\begin{eqnarray} \label{eq_operators_of_FPDE}
	L:=\sum_{k=1}^m{_{0}^{C}D_{t}^{\beta_k}}-\Delta -\int\cdot\  d\boldsymbol{s}, 
\end{eqnarray}
where the operator $\int\cdot\  d\boldsymbol{s}$ denotes a Volterra or Fredholm integral, 
${}_0^C D_t^{\beta_k}$ denotes temporal Caputo fractional derivative 
of order $\beta_k \in (0,1)\cup (1,2)$ for $k=1,2,\ldots,m$, 
with $\beta_1 < \beta_2 < \cdots < \beta_m$. In practice, 
each term ${}_0^C D_t^{\beta_k}$ could be multiplied by a positive constant, 
we set the constant to 1 here for simplicity. Multi-term fractional operators 
enhance the accuracy of modeling complex systems by capturing 
hierarchically distributed memory effects and non-local interactions 
across multiple time scales, enabling precise characterization of anomalous 
diffusion dynamics  and viscoelastic relaxation with power-law memory decay. 
For instance, Srivastava et al. \cite{srivastava2010multi} 
employ multi-term fractional diffusion 
equations to characterize oxygen subdiffusion in anisotropic media, 
while Torvik and Bagley \cite{torvik1984} 
use two-term formulations to discriminate between solute transport regimes. 
A critical challenge in solving (\ref{FPIDE_problem}) arises from their 
inherent non-smoothness in the temporal direction, particularly 
exhibiting initial layer behaviors near $t=0$ \cite{jin2019subdiffusion,she2022transformed,zhang2024fast}. 
To solve this difficulty, 
researchers have developed specialized numerical schemes. For example, She et al. \cite{she2022transformed} introduce a change of variable 
in the temporal direction and use a modified $\mathrm{L1}$ approximation 
for the Caputo derivatives, combined with a Galerkin-spectral method 
for spatial discretization. 
Zeng et al. \cite{zeng2017second} design a modified formula with the 
appropriate correction terms.

Machine learning approaches solving (\ref{FPIDE_problem}) typically 
employ neural networks as trial functions and the residual of the 
equation as the loss function. 
They mainly focus on two aspects: calculation of the loss function, 
and optimization of neural network parameters.
The nonlocal property makes TFPIDEs be high-dimensional integro-differential equations, posing fundamental challenges for efficiently computing the loss function. Among those ingredients, the most crucial one is to effectively approximate the Caputo derivative as well as the Fredholm or Volterra integral. Based on this consideration, we propose a temporal-spatial separated TNN architecture where the TNN function is explicitly multiplied by $ t^{\mu}$, where the exponent $\mu$ is defined in \eqref{eq:def_mu}. Then Gauss-Jacobi quadrature is used to discretize Caputo fractional derivatives. This design achieves two advantages 
for solving equation \eqref{FPIDE_problem}:
\begin{itemize}
	\item When using spatiotemporal-separated TNNs as the trial function, 
	the fractional derivative in the temporal direction can be straightforwardly 
	generated by applying the fractional-order operator to the temporal neural 
	network in conjunction with spatial data, enabling parallel computation 
	of the temporal and spatial derivatives. Additionally, 
	the variable-separated structure of TNN  enables the use of high-density 
	numerical quadrature points in the loss function calculation for high accuracy. 
	Therefore, both the efficiency and accuracy of loss function computation are guaranteed.
	\item The TNN function multiplied by $ t^{\mu}$ inherently satisfies the homogeneous 
	initial condition. Based on careful analysis, we propose a suitable strategy for selecting the exponent $\mu$, which is closely related to the order of the fractional-order operator. 
	This design is beneficial to enhance the effectiveness of the Gauss-Jacobi quadrature without introducing additional singularities into the neural network.
\end{itemize}
Compared with fPINN methods and other machine learning methods, the method 
proposed in this paper shows obvious accuracy improvements 
for solving TFPIDEs (\ref{FPIDE_problem}).

An outline of the paper goes as follows. We introduce in Section 
\ref{section:construct and calculate the loss function} the TNN structure, 
the construction of the loss function, and the numerical schemes of Caputo derivatives.
In Section \ref{section_Optimization}, 
the machine learning method, which is 
designed to obtain the optimal TNN parameters, will be built for solving 
linear and nonlinear TFPIDEs. 
Section \ref{Section_Numerical} provides some numerical examples to validate 
the accuracy and efficiency of the proposed method.

\section{Construction and computation of the loss function for solving TFPIDEs}\label{section:construct and calculate the loss function}
In order to construct and compute the loss function based on TNN for solving TFPIDEs, in this section, we introduce architecture of spatiotemporal-separated TNN, the construction of the loss function, and the discretization processes of the Caputo fractional derivatives.  In addition, the discretizations for Fredholm and Volterra integrals will be introduced in numerical examples.

\subsection{Tensor neural network architecture }\label{Section_TNN}

In this section, we introduce the TNN structure and its approximation properties, 
and some techniques to improve the numerical stability. The approximation property 
and computational complexity of related integration associated with the TNN 
structure have been discussed in \cite{wang2022tensor}.

\begin{figure}[htb]
	\centering
	\includegraphics[width=13cm,height=7.5cm]{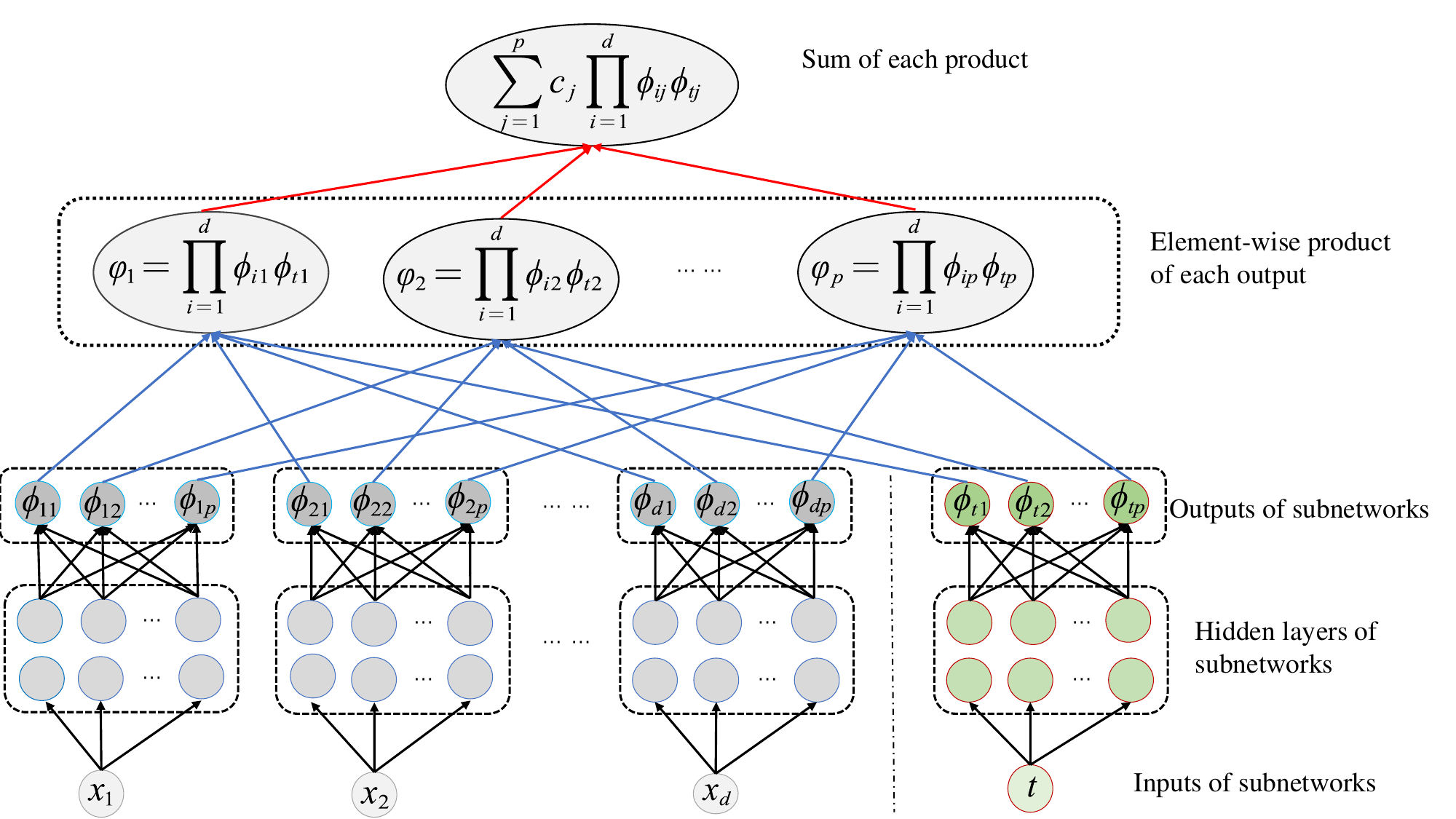}
	\caption{Architecture of spatiotemporal-separated TNN. Black arrows mean 
		linear transformation (or affine transformation). Each ending node of 
		blue arrows is obtained by taking the scalar multiplication of 
		all starting nodes of blue arrows that end in this ending node. 
		The final output of TNN is derived from the summation of all starting nodes 
		of red arrows.}\label{TNNstructure}
\end{figure}

The TNN is built with $d$ subnetworks, and each subnetwork is 
a continuous mapping from a bounded closed set $\Omega_i\subset\mathbb R$ 
to $\mathbb R^p$, which can be expressed as
\begin{eqnarray*}
	\Phi_i(x_i;\theta_i)&=&\big(\phi_{i,1}(x_i;\theta_i), \phi_{i,2}(x_i;\theta_i),\cdots,\phi_{i,p}(x_i;\theta_i)\big)^{\top},\\
	\Phi_t(t;\theta_t)&=&(\phi_{t,1}(t;\theta_t),\phi_{t,2}(t;\theta_t),\cdots ,\phi_{t,p}(t;\theta_t))^{\top}, 
\end{eqnarray*}
where $i=1, \cdots, d$, each $x_i$ denotes the one-dimensional input, 
$\theta_i$ denotes the parameters of the $i$-th subnetwork, 
typically the weights and biases. 
As illustrated in Figure \ref{TNNstructure}, the TNN structure containing 
time is composed of  $p$ fully connected neural networks (FNNs) 
for spatial basis functions $\Phi_i(x_i;\theta_i), i=1,2,\ldots,d$, 
and one FNN for the temporal basis function $\Phi_t(t;\theta_t)$.

In order to improve the numerical stability, we normalize each 
$\phi_{i,j}(x_i)$, $\phi_{t,j}(t)$ and use the following normalized TNN structure:
\begin{eqnarray}\label{def_TNN_normed}
	\Psi (\boldsymbol{x},t;c,\theta )&=&\sum_{j=1}^p{c_j}\widehat{\phi}_{1,j}(x_1;\theta_1)
	\cdots \widehat{\phi}_{d,j}(x_d;\theta_d)\widehat{\phi}_{t,j}(t;\theta_t)\nonumber\\
	&=&\sum_{j=1}^p{c_j}\prod_{i=1}^d{\widehat{\phi }_{i,j}(x_i;\theta_i)
		\widehat{\phi }_{t,j}(t;\theta_t)}
	=:\sum_{j=1}^p{c_j}\varphi_j(\boldsymbol{x},t;\theta), 
\end{eqnarray}
where each $c_j$ is a scaling parameter with respect to the normalized rank-one function, 
$c=\{c_j\}_{j=1}^{p}$ denotes the linear coefficients for the 
basis system built by the rank-one functions $\varphi _j(x,t;\theta )$, $j=1, \cdots, p$. 
For $i=1,\cdots,d$, $j=1,\cdots,p$,  $\widehat\phi_{i,j}(x_i;\theta_i)$ 
and $\widehat{\phi }_{t,j}(t;\theta _t)$ are $L^2$-normalized function as follows:
\begin{eqnarray}\label{normalized}
	\widehat{\phi }_{i,j}(x_i;\theta _i)=\frac{\phi _{i,j}(x_i;\theta _i)}
	{\left\| \phi _{i,j}(x_i;\theta _i) \right\| _{L^2(\Omega _i)}}
	, \quad \widehat{\phi }_{t,j}(t;\theta _t)=\frac{\phi _{t,j}(t;\theta _t)}
	{\left\| \phi _{t,j}(t;\theta _t) \right\| _{L^2(\Omega _i)}}.
\end{eqnarray}
For simplicity of notation, $\phi_{i,j}(x_i;\theta_i)$ 
and $\phi_{t,j}(t;\theta _t)$ denote the normalized function in the following parts.

Due to the isomorphism relation between $L^2(\Omega_1\times\cdots\times\Omega_{d+1})$ 
and the tensor product space $L^2(\Omega_1)\otimes\cdots\otimes L^2(\Omega_{d+1})$ 
with $\Omega_{d+1}:=(0,T]$,  
the process of approximating the function $f(\boldsymbol{x},t)\in L^2(\Omega_1\times\cdots\times\Omega_{d+1})$ 
with the TNN defined by (\ref{def_TNN_normed}) is actually to search 
for a correlated CP decomposition 
to approximate $f(\boldsymbol{x},t)$ in the space $L^2(\Omega_1)\otimes\cdots\otimes L^2(\Omega_{d+1})$ 
with rank not greater than $p$.  Due to the low-rank structure, 
we will find that the polynomial mapping acting on the TNN and its 
derivatives can be done with small 
scale computational work \cite{wang2022tensor}.  To demonstrate the 
effectiveness of solving FPDEs using the TNN method, we introduce the following 
approximation result for functions in the space $H^{m}(\Omega_1\times\cdots\times\Omega_{d+1})$ 
with respect to the $H^m(\Omega_1\times\cdots\times\Omega_{d+1})$-norm, 
where $m$ is a non-negative integer.
\begin{theorem} \cite{wang2022tensor}\label{theorem_approximation}
	Assume that each $\Omega_i$ is a bounded closed interval in $\mathbb R$ 
	for $i=1, \cdots, d$, $\Omega=\Omega_1\times\cdots\times\Omega_{d}$, 
	and the function $f(\boldsymbol{x},t)\in H^m(\Omega \times (0,T])$ with a non-negative integer $m$. 
	Then for any tolerance $\varepsilon>0$, 
	there exists a positive integer $p$ and the corresponding TNN basis functions defined by (\ref{def_TNN_normed}) such that the following approximation property holds
	\begin{equation}\label{eq:L2_app}
		\|f(\boldsymbol{x},t)-\Psi(\boldsymbol{x},t;c,\theta)\|_{H^m(\Omega \times (0,T])}<\varepsilon.
	\end{equation}
\end{theorem}
\begin{remark}
	Note that for the spatial direction, we employ the variable-separated neural network, 
	namely the TNN as well. While the non-variable-separated FNN can also be used, the advantage of using TNN is that the partial derivatives in the spatial direction can be efficiently obtained, 
	and the high-dimensional integration in the loss function can be transformed into one-dimensional integration in some cases. Therefore, both the efficiency and accuracy in the calculation process of the loss function can be guaranteed. Meanwhile, the approximation ability of TNN makes it reasonable as a trial function to solve partial differential equations. In our numerical examples, 
	we also present a problem whose exact solution is not in a variable-separated form, 
	yet the TNN method still achieves high accuracy. 
\end{remark}

\subsection{Construction of the loss function} \label{section:loss_function}
Many studies have shown that an imbalance between PDE loss and initial/boundary loss may lead to a decrease in accuracy and significantly increase the training cost 
even there have appeared weighting techniques to correct this 
imbalance \cite{xu2024subspace,zhang2024adaptive}. 
For this reason, in this section, we provide a method 
so that there is no need to introduce the penalty term to enforce the initial/boundary 
value conditions into the loss function. This can improve the convergence rate 
and stability of the training process.

If $g(\boldsymbol{x},t)=0$ and $s(\boldsymbol{x})=0$ in the problem (\ref{FPIDE_problem}), to ensure that the TNN function $\Psi$ satisfies the boundary conditions, the formula for the left-hand side of (\ref{normalized}) can be modified  as follows:
\begin{eqnarray}\label{bd_condition}
	\widehat{\phi }_{i,j}(x_i;\theta _i)
	=\frac{\phi _{i,j}(x_i;\theta _i)(x_i-a_i)(b_i-x_i)}{\left\| \phi _{i,j}(x_i;\theta _i)(x_i-a_i)(b_i-x_i) \right\| _{L^2(\Omega _i)}},
\end{eqnarray}
where the factor function $(x_i - a_i)(b_i-x_i)$ is used for treating the homogeneous Dirichlet boundary conditions defined on $\partial \Omega$. 
In addition, we multiply the TNN function $\Psi$ by the  function $t^{\mu}$ with $\mu >0$. Therefore, the TNN trial function $\Psi (\boldsymbol{x},t;c,\theta )$ is defined as follows: 
\begin{eqnarray}\label{TNN_trial_fun}
	\Psi (\boldsymbol{x},t;c,\theta )=\sum_{j=1}^p{c_j}\prod_{i=1}^d{\widehat{\phi} _{i,j}(x_i;\theta _i)t^{\mu}\widehat{\phi} _{t,j}(t;\theta _t)}.
\end{eqnarray}

The choice of the exponent $\mu$ significantly influences the properties of the trial function. For different types of TFPIDEs (\ref{FPIDE_problem}), we adopt distinct strategies for selecting $\mu$. 
Specifically, $\mu$ is defined as follows:
\begin{equation}\label{eq:def_mu}
	\mu =\left\{ \begin{array}{ll}
		\beta_{1}, & {\rm if}\ \beta_{m} \in (0,1),\\
		\beta_{i_0},  & 
		{\rm if}\ \beta_{m} \in (1,2),
	\end{array} \right.
\end{equation}
where $\beta_{i_0}:=\mathop {\min} \limits_{i=1,2,..,m}\left\{ \beta _i,\beta _i>1 \right\}$. This selection strategy is motivated by the following considerations. On one hand, multiplying the trial function by $t^\mu$ ($\mu>0$) is essential to ensure that $ \Psi $ 
satisfies the initial condition; on the other hand, the choice of $\mu$ considerably affects the smoothness and behavior of the trial function, and thus requires careful considerations:
\begin{enumerate}
	\renewcommand{\labelenumi}{(\alph{enumi})}
	\item Case $\beta_{m}\in (0,1)$: To avoid introducing additional singularities through the multiplicative factor $t^\mu$ into the neural network, $\mu$ should be no greater than the smallest fractional order, i.e., $0<\mu\leq \beta_1$. 
	\item Case $\beta_{m}\in (1,2)$: To ensure that the time Caputo fractional derivative of order $\beta_m$ of the trial function \eqref{TNN_trial_fun} is well-defined in the sense of Riemann-Liouville fractional integral, $\mu$ must satisfy $\mu \geqslant 1$. Likewise, keeping $\mu \leq \beta_{i_0}$ avoids introducing additional singularities into the neural network. Additionally, choosing $\mu>1$ allows the singular term $\tau^{\mu - 2}$ to be absorbed into the weight function of the Gauss–Jacobi quadrature, thereby enhancing the smoothness of the integrand in \eqref{eq:GJ_s_mu_1_2} and improving numerical integration accuracy. Hence, the appropriate range for $\mu$ in this case is $1< \mu \leq \beta_{i_0}$.
\end{enumerate}
It is worth noting that in \eqref{eq:GJ_s_mu} and \eqref{eq:GJ_s_mu_1_2}, the closer $\mu$ is to zero and one  respectively, the greater smoothness of the integrand requires to obtain high quadrature accuracy, due to the property of Gauss-Jacobi quadrature. This aspect should also be taken into consideration when determining an appropriate value for $\mu$. Overall, selecting the optimal exponent $\mu$ within the appropriate range discussed above is problem-dependent. In this work, we adopt the definition in \eqref{eq:def_mu}, which is also consistent with the analysis in \cite{GU2021110231}. 

With trial function \eqref{TNN_trial_fun}, 
the corresponding loss function is given by
\begin{eqnarray}\label{section:loss_function_0}
	\mathcal{L}=\left\|L\Psi(\boldsymbol{x},t;c,\theta)
	-f(\boldsymbol{x},t,\Psi(\boldsymbol{x},t;c,\theta))\right\|_{\Omega \times (0,T]}.
\end{eqnarray}

For the non-homogeneous initial boundary value conditions, 
we introduce a new function $\widehat{g}(\boldsymbol{x},t)\in H^1(\Omega \times (0,T])$ 
such that 
$$\widehat{g}(\boldsymbol{x},t)=g(\boldsymbol{x},t ), 
\ \ ( \boldsymbol{x},t )\in \partial\Omega \times (0,T].$$ 
Then using the formula, 
\begin{equation}
	u(\boldsymbol{x}, t) = \widehat{u}(\boldsymbol{x},t) + s(\boldsymbol{x})-g(\boldsymbol{x},0) +\widehat{g}(\boldsymbol{x},t), 
\end{equation}
the non-homogeneous initial boundary value problem is transformed into the corresponding homogeneous one. 
For more information about the treatment for nonhomogeneous boundary value conditions, please refer to \cite{wang2024solving}.

\subsection{Compute the Caputo fractional derivative}\label{section:evaluation of Caputo and integrals}
In this subsection, we use the Gauss-Jacobi quadrature scheme to numerically approximate Caputo fractional derivatives.  The Caputo fractional derivative \cite{jiang2017fast} of order $\nu >0$ for the given function $f(t)$, $t\in(0,T]$ is defined as follows:
\begin{eqnarray}\label{eq:def_of_caputo_fra}
	_{0}^{C}\mathrm{D}_{t}^{\nu}f(t)=\frac{1}{\Gamma (n-\nu )}
	\int_0^t{(t}-s)^{n-\nu -1}f^{(n)}(s)\mathrm{d}s,
\end{eqnarray}
where $\Gamma$ denotes the Gamma function and $n $ is a positive 
integer satisfying $n-1 < \nu < n$.

The Gauss-Jacobi numerical integration is an efficient method for calculating 
definite integrals with high accuracy, it uses $N$ nodes and weights to approximate 
the integral over $[-1,1]$ with a weighting function of 
$(1 - t)^{\alpha}(1+t)^{\beta}$ (cf. \cite{brzezinski2018computation})
\begin{equation}\label{Def_gauss_jacobi_11}
	\int_{-1}^1{(1}-t)^{\alpha}(1+t)^{\beta}f(t)dt=\sum_{i=1}^{N}{\widehat{w}_{i}^{(\alpha ,\beta )}f(\widehat{t}_{i}^{(\alpha ,\beta )})}+R_{N}(f),
\end{equation}
where $\alpha >-1$, $\beta >-1$, the $N$ nodes 
$\{\widehat{t}_{i}^{(\alpha,\beta)}\}_{i=1}^{N}$ are 
the zeros of the Jacobi polynomial of degree $N$, $P_{N}^{(\alpha, \beta)}(t)$.
And the $N$ weights $\{\widehat{w}_{i}^{(\alpha,\beta)}\}_{i=1}^{N}$ 
can be calculated by the following formula,
\begin{eqnarray*}
	\widehat{w}_{i}^{(\alpha ,\beta )}=2^{\alpha +\beta +1}
	\frac{\Gamma (\alpha +N+1)\Gamma (\beta +N+1)}{N!\Gamma (\alpha +\beta +N+1)
		\left( 1-\left( \widehat{t}_{i}^{(\alpha ,\beta )} \right) ^2 \right) 
		\left[ P_{N}^{(\alpha ,\beta )\prime}
		\left( \widehat{t}_{i}^{(\alpha ,\beta )} \right) \right] ^2}.
\end{eqnarray*}
About the integration error $R_N(f)$  and more information, please refer to  \cite{jahanshahi2017fractional}.   

We design the Gauss-Jacobi quadrature scheme on the interval $[0, 1]$ as follows
\begin{eqnarray}\label{Gauss_Jacobi_2}
	\int_0^1{(}1-t)^{\alpha}t^{\beta}f(t)dt\approx \sum_{i=1}^N{w_{i}^{(\alpha ,\beta )}}f(t_{i}^{(\alpha,\beta )}),
\end{eqnarray}
where we make affine transformations $t_{i}^{(\alpha ,\beta )}
=\frac{1}{2}\widehat{t}_{i}^{(\alpha ,\beta )}+\frac{1}{2}$ 
and $w_{i}^{(\alpha,\beta )}=\frac{1}{2^{\alpha +\beta +1}}\widehat{w}_{i}^{(\alpha,\beta )}$.

For simplicity of notation, $\phi_{i, j} (\cdot) $ and $\phi_{t, j} (\cdot) $ 
denote $\phi_{i, j} (\cdot, \theta_{i}) $ and $\phi_{t, j} (\cdot, \theta_{t})$, respectively.
Thus, for the Caputo derivative of order $\nu \in (0,1)$, 
we have the following computing scheme 
\begin{eqnarray}
	_{0}^{C}D_{t}^{\nu}t^{\mu}\phi _{t,j}(t)&=&\frac{1}{\Gamma (1-\nu )}\int_0^t{(t}-s)^{-\nu}\frac{\partial s^{\mu}\phi _{t,j}(s)}{\partial s}\mathrm{d}s \nonumber \\ \label{eq:GJ_s_mu}
	(\mathrm{let}\,s=t\tau )&=&\frac{1}{\Gamma (1-\nu )}t^{1-\nu}\int_0^1{(1-\tau )^{-\nu}\tau ^{\mu -1}S_1\left( t,\tau \right) d\tau}\\
	&\approx& \frac{1}{\Gamma (1-\nu )}\sum_{k=1}^{N_{\tau}}{w_{k}^{(-\nu ,\mu -1)}t^{1-\nu}S_1( t,\tau _{k}^{(-\nu ,\mu -1)} )}, \nonumber
\end{eqnarray}
where the function $S_1\left( t,\tau \right) =\mu t^{\mu -1}\phi _{t,j}(t\tau )+t^{\mu}\tau \phi _{t,j}^{'} (t\tau )$.

Similarly, when the fractional order $\nu \in (1,2)$, we have the following computing scheme
\begin{eqnarray}
	_{0}^{C}D_{t}^{\nu}t^{\mu}\phi _{t,j}(t)&=&\frac{1}{\Gamma (2-\nu )}\int_0^t{(t}-s)^{1-\nu}\frac{\partial ^2s^{\mu}\phi _{t,j}(s)}{\partial s^2}\mathrm{d}s \nonumber \\ \label{eq:GJ_s_mu_1_2}
	(\mathrm{let}\,s=t\tau )&=&\frac{1}{\Gamma (2-\nu )}t^{2-\nu}\int_0^1{(1}-\tau )^{1-\nu}\tau ^{\mu -2}S_{2}\left( t,\tau \right) d\tau \\
	&\approx& \frac{1}{\Gamma (2-\nu )}\sum_{k=1}^{N_{\tau}}{w_{k}^{(1-\nu ,\mu -2)}t^{2-\nu}S_{2}(t,\tau _{k}^{(1-\nu ,\mu -2)})}, \nonumber
\end{eqnarray}
where the function $S_{2}\left( t,\tau \right) = \mu (\mu -1)t^{\mu -2}\phi _{t,j}(t\tau )+2\mu t^{\mu -1}\tau \phi _{t,j}^{'} (t\tau )+t^{\mu}\tau^{2} \phi _{t,j}'' (t\tau ) $.

Applying the fractional derivative operator $_{0}^{C}\mathrm{D}_{t}^{\nu}$ to 
the TNN function \eqref{TNN_trial_fun} leads to 
\begin{eqnarray*}
	_{0}^{C}\mathrm{D}_{t}^{\nu}\Psi (\boldsymbol{x},t;c,\theta )= 
	\sum_{j=1}^p{c_j\prod_{i=1}^d{\phi_{i,j}(x_i;\theta_i)}_{0}^{C}D_{t}^{\nu}t^{\mu}\phi_{t,j}(t;\theta_{t})}.
\end{eqnarray*}

Note that MC-fPINNs \cite{guo2022monte} and Improved MC-fPINNs \cite{hu2024tackling} transform the Caputo derivative \eqref{eq:def_of_caputo_fra} with $0<\nu<1$ by the following scheme
\begin{equation}\label{FD_Caputo}
	_{0}^{C}D_{t}^{\nu}f(t)=\frac{1}{\Gamma \left( 1-\nu \right)}\left[ \nu t^{1-\nu}\int_0^1{\tau ^{-\nu}\frac{f(t)-f(t-t\tau )}{t\tau}d\tau}+\frac{f(t)-f(0)}{t^{\nu}} \right].
\end{equation}
Then MC-fPINNs evaluates the integral \eqref{FD_Caputo} using Monte Carlo (MC) method while
Improved MC-fPINNs uses Gauss-Jacobi quadrature instead.
As noted in \cite{hu2024tackling}, one of the advantages of using Gauss-Jacobi quadrature over MC method for evaluating \eqref{FD_Caputo} or \eqref{eq:GJ_s_mu} in this article is that fixed quadrature points can therefore be used, eliminating the need for resampling during the training process. Another benefit lies in its treatment of the weak singularity. Specifically, Gauss-Jacobi quadrature incorporates the singular factor in the integrand as part of the weight function, thereby effectively removing the singularity from the numerical integration and achieving higher accuracy than MC method. In contrast, the MC method used in MC-fPINN struggles to accurately handle the weak singularity inherent in the definition of the time-fractional derivative \eqref{eq:def_of_caputo_fra}.

\section{TNN-based machine learning method for TFPIDEs} \label{section_Optimization} 
This section is devoted to introducing the machine learning method, which is designed to obtain the optimal TNN parameters, for solving TFPIDEs with high accuracy. The basic idea is to generate a subspace using the output functions
of TNN as the basis functions, where we find an optimal approximate solution in the least-squares sense.
Then the training steps are adopted to update the TNN subspace to enhance its approximation capability.

We define the TNN subspace as follows: 
\begin{equation*}
	V_{p}:={\rm span}\Big\{\varphi_{j}(\boldsymbol{x},t;\theta), j=1, \cdots, p\Big\},
\end{equation*}
where $\varphi_{j}(\boldsymbol{x},t;\theta)$ is the rank-one function defined in (\ref{def_TNN_normed}).
Then the subspace approximation procedure can be adopted to get the solution $u_p\in V_p$. 
For this purpose, in this paper, the subspace approximation $u_p\in V_p$ for (\ref{FPIDE_problem}) 
is obtained by minimizing the residual $Lu_p-f$ in the sense of $L^2(\Omega\times (0, T])$. 
Then the corresponding discrete problem can be given as: 
Find $u_p\in V_p$ such that
\begin{eqnarray}\label{eq:least_square_method}
	(Lu_p,Lv_p)=(f,Lv_p),\ \  \forall v_p\in V_p.
\end{eqnarray}
After obtaining the approximation $u_p$, the subspace $V_p$ is updated by optimizing the loss function.

\subsection{Solving linear TFPIDEs}
In this subsection, we introduce the TNN subspace method to solve linear TFPIDEs.
After the $\ell$-th training step, the TNN $\Psi (\boldsymbol{x},t;c,\theta^{(\ell)}) $ in \eqref{TNN_trial_fun} belongs to the following subspace:
\begin{equation*}
	V_{p}^{(\ell )}:=\mathrm{span}\left\{ \varphi _j(\boldsymbol{x},t;\theta ^{(\ell )}), j=1,\cdots ,p \right\},
\end{equation*}
where $\varphi_{j}(\boldsymbol{x},t;\theta ^{(\ell )})$ is the rank-one function defined in (\ref{def_TNN_normed}) after $\ell$ training steps.
\begin{algorithm}[htb!]
	\caption{TNN subspace method for linear TFPIDEs}\label{Algorithm_1}
	\begin{enumerate}
		\item Initialization:
		Build the initial TNN $\Psi(\boldsymbol{x},t;c^{(0)},\theta^{(0)})$ defined in \eqref{TNN_trial_fun}, set the loss function $\mathcal{L}(\Psi (\boldsymbol{x},t;c,\theta) )$ defined in \eqref{section:loss_function_0}, 
		the maximum number of training steps $M$, 
		learning rate $\gamma$, and the iteration counter $\ell = 0$.
		\item Define the subspace $V_{p}^{(\ell)}$ as follows:
		\begin{eqnarray*}
			V_{p}^{(\ell )}:=\mathrm{span}\left\{ \varphi _j(\boldsymbol{x},t;\theta ^{(\ell )}), j=1,\cdots ,p \right\} .
		\end{eqnarray*}
		Assemble the stiffness matrix $A^{(\ell)} \in \mathbb{R}^{p \times p}$ and right-hand side term
		$B^{(\ell)} \in \mathbb{R}^{p}$ on $V_{p}^{(\ell)}$:
		\begin{eqnarray*}
			A_{m,n}^{(\ell )}=(L\varphi _{n}^{(\ell )},L\varphi _{m}^{(\ell)}), \ \ B_{m}^{(\ell )}=(f,L\varphi _{m}^{(\ell )})\ \  1\le m,n\le p.
		\end{eqnarray*}
		\item Solve the following linear system to determine the coefficient vector $c \in \mathbb{R}^{p}$
		\begin{eqnarray*}
			A^{(\ell)}c = B^{(\ell)}.
		\end{eqnarray*}
		Update $\Psi(\boldsymbol{x},t;c^{(\ell+1)},\theta^{(\ell)})$  with the coefficient vector: $c^{(\ell+1)} = c$.
		\item Update the network parameters from $\theta^{(\ell)}$ to $\theta^{(\ell+1)}$, by optimizing the loss function $\mathcal{L}^{(\ell+1)}(c^{(\ell + 1)},\theta ^{(\ell)})$ through gradient-based training steps.
		\item Iteration: set $\ell = \ell + 1$. If $\ell < M$, go to Step 2. Otherwise, terminate.
	\end{enumerate}
\end{algorithm}

According to the definition of TNN and the corresponding space $V_{p}^{(\ell )}$, it is easy to know that the neural network parameters $\theta^{(\ell)}$ determine the space $V_p^{(\ell)}$, 
while the coefficient parameters $c^{(\ell)}$ determine the direction of TNN 
in the space $V_p^{(\ell)}$. 
We can divide the optimization step into two sub-steps. Firstly, 
assume the neural network parameters $\theta^{(\ell)}$ are fixed. The optimal coefficient parameters $c^{(\ell+1)}$ 
can be obtained with the least squares method (\ref{eq:least_square_method}), i.e., 
by solving the following linear equation 
\begin{eqnarray}\label{linear_system}
	A^{(\ell)}c^{(\ell+1)}=B^{(\ell)},
\end{eqnarray}
where the matrix $A^{(\ell)}\in \mathbb R^{p\times p}$ and the vector 
$B^{(\ell)}\in\mathbb R^{p\times 1}$ are assembled as follows: 
\begin{eqnarray*}
	A_{m,n}^{(\ell )}=(L\varphi _{n}^{(\ell )},L\varphi _{m}^{(\ell )} ),\ \  
	B_{m}^{(\ell )}=(f,L\varphi _{m}^{(\ell )}), \ \  1\le m,n\le p.
\end{eqnarray*}
Secondly, when the coefficient parameters $c^{(\ell+1)}$ are fixed, 
the neural network parameters $\theta^{(\ell+1)}$ can be updated by 
the optimization steps included, such as Adam or L-BFGS for the loss function $\mathcal{L}^{(\ell+1)}(c^{(\ell + 1)},\theta ^{(\ell)})$ defined in Subsection \ref{section:loss_function}.
Thus, the TNN-based machine learning method for the homogeneous Dirichlet 
boundary and initial value problem can be defined in Algorithm \ref{Algorithm_1}. 
From the perspective of optimization, Algorithm \ref{Algorithm_1} can be interpreted as an alternating optimization procedure for the unknown parameters $c$ and $\theta$ in the trial function \eqref{TNN_trial_fun}, where the linear coefficients $c$ are optimally obtained in the least-squares sense, and the neural network parameters $\theta$ are updated by minimizing the loss function. Numerical results show that the alternating optimization procedure has reduced the difficulty of the optimization problem, and the convergence of the training process is significantly improved.

Note that the use of the Gauss-Jacobi quadrature scheme, as introduced in Subsection \ref{section:evaluation of Caputo and integrals}, ensures high accuracy 
in discretizing the Caputo fractional derivative. Meanwhile, the variable-separated 
structure of TNN transforms the integration in the loss function into one-dimensional 
integration, which can be effectively and accurately calculated using numerical 
quadrature scheme. Due to the same reason, all integrations involved in Algorithm \ref{Algorithm_1} 
can be computed efficiently with high accuracy. It's well known that the optimization error 
and the integration error are usually two primary sources of overall 
error in machine learning methods for solving PDEs; 
see, for example, \cite{mishra2021estimates}. 
The TNN subspace-based method proposed in this paper explicitly takes 
both sources of error into account, thereby demonstrating strong potential 
for achieving high accuracy in solving TFPIDEs.

\subsection{Solving nonlinear TFPIDEs} \label{section:optimization for nonlinear}
This subsection is devoted to designing a TNN-based machine learning method for solving nonlinear problems (\ref{FPIDE_problem}). Here,  the operator of the problem (\ref{FPIDE_problem}) with respect to the unknown solution $u$ is assumed to be expressed as $L_1 + G$, where $L_1$ is a linear operator and $G$ is a nonlinear operator. Specifically, in this paper,  we consider the following two cases: 
\begin{eqnarray}
	&&G(u)(\boldsymbol{x},t)=u^2(\boldsymbol{x},t), \label{eq:Nonlinear_Case_1}\\
	&&G(u)(t) =\int{k(\boldsymbol{s},t)u^2(\boldsymbol{s},t)}d\boldsymbol{s}. \label{eq:Nonlinear_Case_2}
\end{eqnarray}

We denote the solution at the $k$-th outer iteration and $\ell$-th 
inner iteration as $u^{(k,\ell)}$ 
\begin{eqnarray*}\label{eq_u_def}
	u^{(k,\ell)} = \Psi(\boldsymbol{x},t;c^{(\ell)},\theta^{(k)}).
\end{eqnarray*}

The iterative method for solving nonlinear problem (\ref{FPIDE_problem}) consists of 
outer and inner iteration steps. In the outer iteration step, the coefficient $c^{(\ell)}$ is fixed, and only the network parameters $\theta^{(k)}$ are updated by the loss function. When the inner iteration is performed, the network parameters $\theta^{(k)}$ are fixed, and the optimal coefficient parameters $c^{(\ell+1)}$ can be obtained by using the fixed-point iteration. Of course,  other types of nonlinear iteration methods can also be used here. 

Firstly, we introduce the inner iteration process. 
Starting from the current solution $u^{(k,\ell)}$, the next step 
in the inner iteration involves solving the following linear equation:
\begin{eqnarray}\label{Linear system}
	L_{1}u^{(k,\ell +1)}+\widehat{G}(u^{(k,\ell )},u^{(k,\ell +1)})=h,
\end{eqnarray}
where $h$ is defined in (\ref{FPIDE_problem}) and $\widehat{G}(u^{(k,\ell )},u^{(k,\ell +1)})$ 
is defined as follows
\begin{eqnarray}
	&&\widehat{G}(u^{(k,\ell )},u^{(k,\ell +1)})=u^{(k,\ell )}u^{(k,\ell +1)}, \label{eq:Nonlinear_Case_1_hat}\\
	&&\widehat{G}(u^{(k,\ell )},u^{(k,\ell +1)})=\int{k(\boldsymbol{s},t)u^{(k,\ell )}(\boldsymbol{s},t)}u^{(k,\ell +1)}(\boldsymbol{s},t)d\boldsymbol{s}.\label{eq:Nonlinear_Case_2_hat}
\end{eqnarray}

In order to solve the linear system (\ref{Linear system}), we assemble 
the stiffness matrix $A^{(\ell+1)}$ and the right-hand side term $B^{(\ell+1)}$ as follows 
\begin{eqnarray*}
	&&A_{m,n}^{(\ell+1)}=(F\varphi _m(\boldsymbol{x},t;\theta ^{(k)}),F\varphi _n(\boldsymbol{x},t;\theta ^{(k)})), \ \ 1\leqslant m, n\leqslant p,\\
	&&B_{m}^{(\ell+1)}=(F\varphi _m(\boldsymbol{x},t;\theta^{(k)}),h) ,\ \ 1\leqslant m\leqslant p,
\end{eqnarray*}
where the term $F\varphi _m(\boldsymbol{x},t,\theta ^{(k)})$ is defined as follows 
\begin{eqnarray*}
	F\varphi _m(\boldsymbol{x},t;\theta ^{(k)}):=L_1\varphi _m(\boldsymbol{x},t;\theta ^{(k)})
	+\sum_{j=1}^p{c_{j}^{(\ell )}}\widehat{G}(\varphi_j(\boldsymbol{x},t;\theta ^{(k)}),\varphi _m(\boldsymbol{x},t;\theta ^{(k)}),\label{eq:nonlinear_operator}
\end{eqnarray*} 
where $\widehat{G}(\cdot,\cdot )$ is defined in (\ref{eq:Nonlinear_Case_1_hat}) or (\ref{eq:Nonlinear_Case_2_hat}).
Then solve the linear system $A^{(\ell+1)}c=B^{(\ell+1)}$ and update the 
new coefficient vector $c^{(\ell+1)} = c$.

After completing the inner iterations, we obtain  the solution $u^{(k,M_2)}$. 
To initiate the outer iteration, we define a loss function as follows:
\begin{equation*}
	\mathcal{L}^{(k+1)}(c^{(M_2)},\theta ^{(k)})
	:=\left\| L_1u^{\left( k,M_2 \right)}-G(u^{\left( k,M_2 \right)})
	-h \right\|_{\Omega \times (0,T]},\label{eq:nonlinear_loss}
\end{equation*}
where $G(\cdot)$ is defined in (\ref{eq:Nonlinear_Case_1}) or (\ref{eq:Nonlinear_Case_2}). 
For the detailed information about the construction of the loss function, please see Subsection \ref{section:loss_function}. 
Then update the network parameters $\theta^{(k)}$ to obtain $\theta^{(k+1)}$ by the optimization steps included, 
such as Adam or L-BFGS for the loss function $\mathcal{L}^{(k+1)}(c^{(M_2)},\theta^{(k)})$. 

As evident from the above description, the overall solution process is almost the same as that for solving linear TFPIDEs, 
with the only difference lying in the way  of obtaining the parameter $c^{\ell+1}$. 
For simplicity, we do not state the corresponding algorithm for solving nonlinear TFPIDEs. 

\section{Numerical examples}\label{Section_Numerical}
In this section, we provide several examples to validate the efficiency and accuracy of 
the proposed TNN-based machine learning method.  All the numerical experiments are done on NVIDIA Tesla A800 GPU and Tesla NVIDIA V100 GPU. 

Relative $L^2$ error at the test points about the approximate solution $\Psi (\boldsymbol{x},t;c,\theta)$ and the exact solution $u$ 
are used to measure the convergence behavior and accuracy of the examples in this section as follows
$$e_{\rm test}:=\frac{\sqrt{\sum_{k=1}^K{( \Psi (x^k,t^k;c,\theta) -u( x^k, t^k))^2}}}{\sqrt{\sum_{k=1}^K{( u( x^k, t^k))^2}}},$$
where the selected test points $ \{(x^k, t^k)\} $ on $  \Omega \times (0, T] $ are defined on a uniform grid of 300 spatial points and 300 temporal points  for one-dimensional FPDEs. For the three-dimensional case in Subsection \ref{example:3d_Volterra}, the test points are defined on a  uniform grid of $ 30^3 $ spatial points and 30 temporal points.

For all examples, the number of Jacobi quadrature nodes is set to 100, each FNN has three hidden layers with 50 neurons and the Tanh as activation function in each layer,  the rank parameter for the TNN is set to be $p = 50$, and set the TNN function such that it can automatically satisfy the initial/boundary value conditions to approximate $\hat{u}(\boldsymbol{x},t)$. 

For one-dimensional examples, during computing the loss function, the interval along the $x$- or $t$-axis is subdivided into 25 subintervals with 16 Gauss points selected within each subinterval for building the quadrature scheme. The TNN is optimized using the Adam optimizer with an initial learning rate of 0.003 and automatically adjusts the learning rate for the 5000 epochs. 
.


\subsection{Linear time-fractional diffusion-wave equations} \label{fractional diffusion-wave equation}
We consider in this subsection the one-dimensional linear time-fractional diffusion wave equation: Find $u(x,t)$ such that
\begin{equation}\label{eq:multi_diffusion_wave_equations}
	\left\{ \begin{aligned}
		\sum_{k=1}^{m}{_{0}^{C}D_{t}^{\beta_k}u(x,t)}&=\frac{\partial^2u(x,t)}{\partial x^2}
		+f(x,t),\ \ (x,t)\in \Omega \times (0,T],\\
		u(x,t)&=0,\ \  (x,t)\in \partial \Omega \times (0,T],\\
		u(x,0)&=s(x),\ \  x\in \Omega. 
	\end{aligned} \right. 
\end{equation}

This is the simplest case of the general problem (\ref{FPIDE_problem}), the interested readers are referred to \cite{gorenflo2015time,huang2013two,jiang2017fast,jiang2011high,ma2023pmnn,ren2023class,
	she2022transformed,srivastava2010multi,torvik1984,zhang2024fast}. 
Gorenflo et al. \cite{gorenflo2015time} proved maximal regularity for equation \eqref{eq:multi_diffusion_wave_equations} in fractional Sobolev spaces. The performance of the proposed TNN-based machine learning method for solving \eqref{eq:multi_diffusion_wave_equations} with the form $t^\alpha$ ($\alpha > 0$) as the exact solution is investigated in this subsection. The function $t^{\alpha}$ with $\alpha \in (0,1)$ has a weak singularity at $t=0$, whose severity intensifies as $\alpha \to 0^+$. In the following examples, we take $\Omega=(0,1)$ and $T=1$.

\subsubsection{Single-term time-fractional diffusion-wave equations}
\label{section:single-term FPDEs}
ln this subsection, we consider the single-term time-fractional diffusion wave equation \eqref{eq:multi_diffusion_wave_equations}.

In the first numerical example, we set the exact solution as 
$u(x,t)=(t^{\alpha_2}+t^{\alpha_1})\sin(2\pi x)$ with $\alpha_1\leqslant \alpha_2$, 
$\alpha_1, \alpha_2 \in (0,4)$, $\beta < \alpha_{1} + 0.5$. 
For this aim, the initial value condition should be $s(x)=0$ and the source term
\begin{eqnarray}\label{single_source_term}
	f(x,t)=\left[\sum_{i=1}^2{\frac{\Gamma(\alpha_i+1)t^{\alpha_i-\beta}}
		{\Gamma (1-\beta +\alpha_i)}}+4\pi^2(t^{\alpha_2}+t^{\alpha_1})\right]\sin(2\pi x).
\end{eqnarray} 

\begin{table}[htb]
	\centering
	\caption{Errors of equation \eqref{eq:multi_diffusion_wave_equations} with $m = 1$ for different values of $\beta,\alpha_1,\alpha_2$ using TNN ($\mu =\beta$).}\label{ex1:tab1}
	\begin{tabular}{cccccc} 
		\toprule
		\multicolumn{3}{c}{$\beta \in (0,1)$} & \multicolumn{3}{c}{$\beta \in (1,2)$} \\ 
		\cmidrule(lr){1-3} \cmidrule(lr){4-6}
		$\beta$ &  $(\alpha_1,\alpha_2)$  &  $e_{\rm test}$ &  $\beta$   & $(\alpha_1,\alpha_2)$ & $e_{\rm test}$ \\ 
		\midrule
		0.01    & (0.01, 0.01)  &  6.395e-08 &  1.10    & (1.20, 1.40)  &  5.405e-07\\
		0.10    & (0.15, 0.20)  &  1.352e-05 &  1.50    & (1.60, 1.70)  &  6.635e-07\\
		\cdashline{4-6}[1pt/1pt] 
		0.01    & (0.20, 0.30)  &  1.537e-05 &  1.30   & (1.20, 1.40)  &  3.001e-06\\
		0.20    & (0.80, 0.90)  &  6.017e-06 &  1.65   & (1.60, 1.80)  &  5.130e-07\\
		\cdashline{1-6}[1pt/1pt]  
		0.50   & (0.01, 0.01)  &  2.067e-04 &  1.40    & (2.00, 2.40)  &  1.707e-07\\ 	
		0.40    & (0.15, 0.20)  &  7.618e-05 &  1.70    & (2.10, 2.90)  &  7.389e-07\\	 	
		\cdashline{1-6}[1pt/1pt] 
		0.50    & (1.40, 1.60)  &  7.350e-07 &  1.40    & (3.00, 3.40)  &  7.412e-08\\  	
		0.60    & (2.30, 2.80)  &  4.362e-07 &  1.70    & (3.50, 3.80)  &  2.307e-08\\
		\bottomrule
	\end{tabular}
\end{table}
\begin{figure}[htb!]
	\centering
	\includegraphics[width=11.8cm,height=4.55cm]{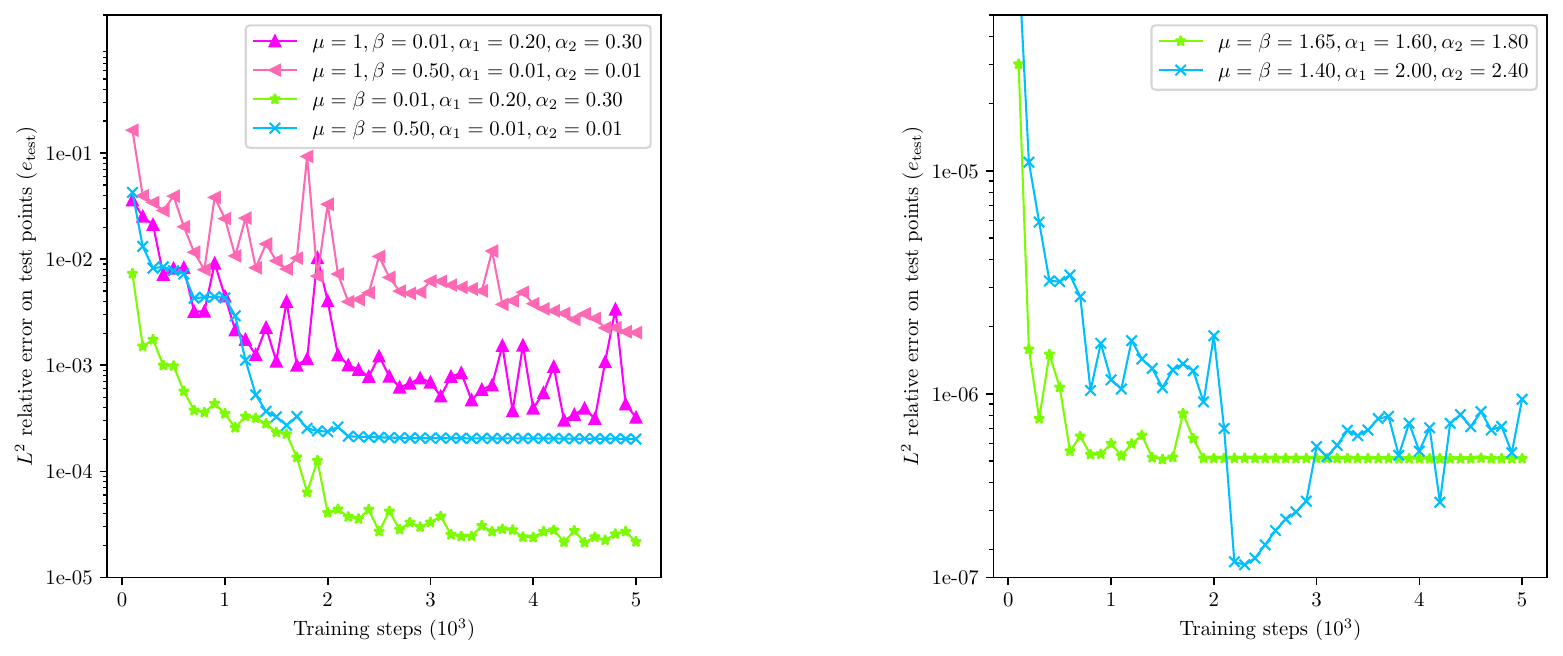}
	\caption{The relative $L^2$ error during the
		training process for solving \eqref{eq:multi_diffusion_wave_equations} with $m = 1$, $\beta \in(0,1)\cup(1,2)$ and  $u=(t^{\alpha_{1}}+t^{\alpha_{2}}) \sin(2\pi x)$.}\label{fig:linear_single_error}
\end{figure}
\begin{figure}[htb!]
	\centering
	\includegraphics[width=13cm,height=3.46cm]{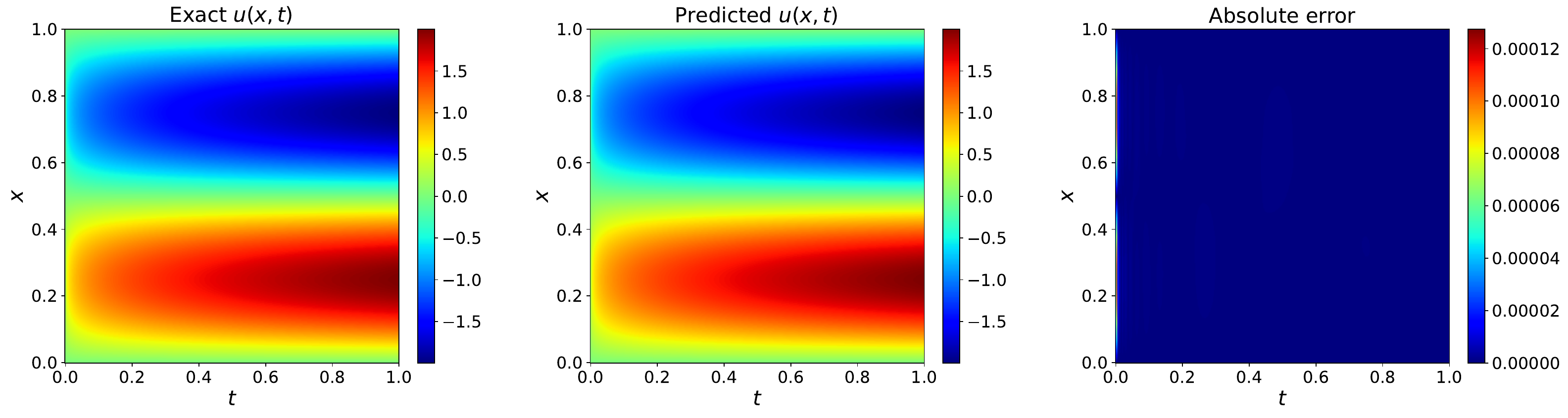}
	\includegraphics[width=12.9cm,height=3.46cm]{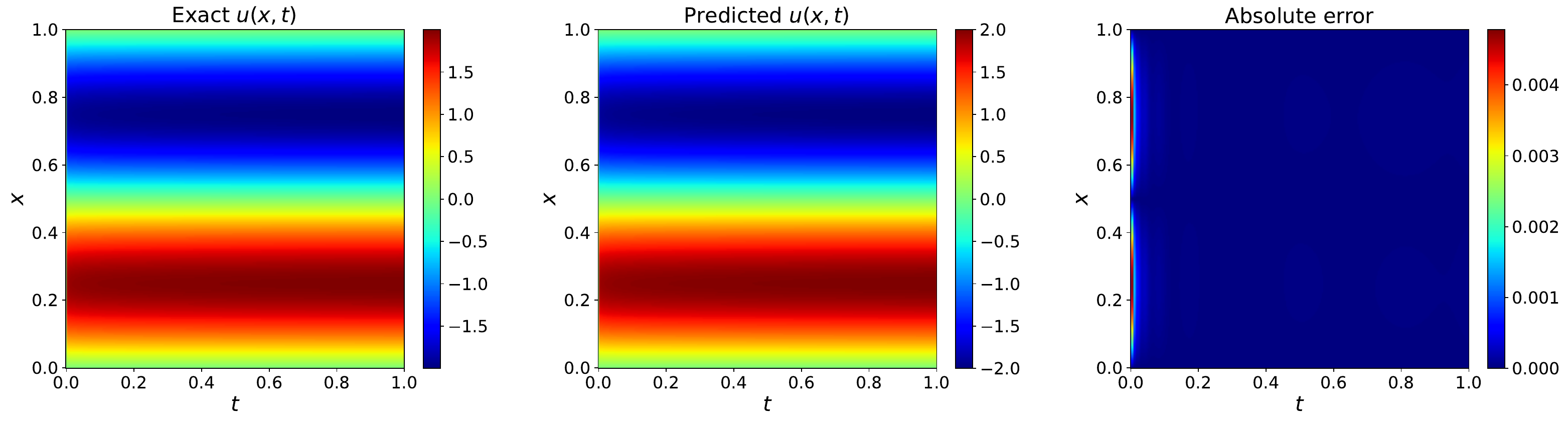}
	\includegraphics[width=12.8cm,height=3.46cm]{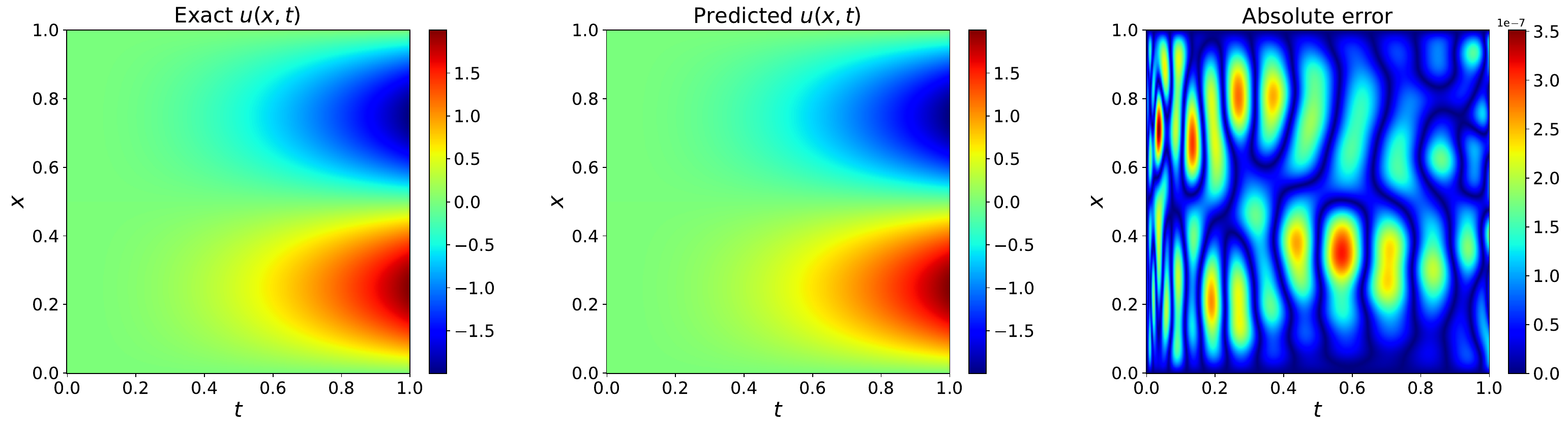}
	\caption{Numerical results of equation \eqref{eq:multi_diffusion_wave_equations} 
		with $m = 1$, 
		$u=(t^{\alpha_{1}}+t^{\alpha_{2}}) \sin(2\pi x)$ 
		for $\alpha_{1} = 0.20, \alpha_{2} = 0.30$, 
		and $\beta =0.01$ (top row); $\alpha_{1} = 0.01,\alpha_{2} = 0.01$, 
		and $\beta =0.5$ (middle row); $\alpha_{1} = 1.40,\alpha_{2} = 2.0$, 
		and $\beta =2.4$ (bottom row): the exact solutions (left column), 
		the predictions (middle column), and the absolute errors of approximation (right column).}\label{fig:linear_single_heat_map} 
\end{figure}

The corresponding numerical results with $\beta \in (0,1)\cup(1,2)$ are 
presented in Table \ref{ex1:tab1}, which shows that the relative 
error attains  around $1 \times 10^{-5}$  for $\beta \in (0,1)$ 
when $\alpha_{1}, \alpha_{2} \in (0, 1) $ and $\alpha_{1}, \alpha_{2}  > \beta $. 
However, the relative error degrades slightly when $\alpha_{1}$ 
or $\alpha_{2} $ are less than $\beta$. 
The left side of Figure \ref{fig:linear_single_error} 
shows that the relative error is significantly improved 
when choosing $\mu=\beta$ rather than  $\mu=1$. 
The top and middle of Figure \ref{fig:linear_single_heat_map} 
show that the absolute error is 
relatively large at time close to $t=0$ when $\alpha_{1}=\alpha_{2}=0.01$ 
and selecting $\mu=\beta$. This is the worst-case, but it still achieves 
a relative error of $2.067 \times 10^{-4}$. 
When $\alpha_{1}, \alpha_{2} \in (1,3)$, 
the function's smoothness improves, which leads to the relative 
error run up to $1 \times 10^{-7}$. 
Similarly, there is a slight decrease in relative error for $\beta \in (1,2) $ and $\alpha_{1},\alpha_{2}<\beta$. 
If $\alpha_{1}, \alpha_{2} \in (2,3)$, the function becomes smooth 
and the relative error gets about $1 \times 10^{-7}$. 
Furthermore, the relative error can arrive 
at about $1 \times 10^{-8}$ when $\alpha_{1}, \alpha_{2} \in (3, 4)$.
As we see, the accuracy of solving the single-term time fractional diffusion-wave equation increases with the smoothness of the exact solution. Specifically, it is related to the value of $\alpha_1- \beta$.
The reason should be that the neural networks are relatively easy to approximate smooth functions and not so easy to catch the strong singularity of the solution.

Additionally, when $\beta \in (0,1)$ and $\beta > \alpha_1$, the fPINN method fails to solve this problem, whereas the TNN method achieves a relative error of around $1 \times 10^{-5}$. 
When $\beta \in (0,1)$ and $\beta < \alpha_1$, the relative error of the fPINN method 
ranges between $1 \times 10^{-2}$ and $1 \times 10^{-3}$, with the error improving 
as the smoothness of the solution increases. 
Notably, when $\beta = \alpha_1 = \alpha_2 = 0.01$, 
the relative error of fPINN is only $1 \times 10^{-1}$. 
These results indicate that fPINN is relatively difficult to solve this problem.

In the second numerical example, we investigate the performance of the TNN-based machine learning method and compare it with other types of machine learning methods.
For $\beta\in (0,1)$, the exact solution of \eqref{eq:multi_diffusion_wave_equations} 
with $m=1$ is set to be $u(x,t)=(t^{\beta}+1) \sin(6\pi x)$, as in \cite{ren2023class} 
with the initial value  condition $s(x) = \sin (6\pi x) $ and the source term
$f(x,t) =\left[ \Gamma (\beta +1 ) +36\pi ^2(t^{\beta}+1 ) \right] \sin(6\pi x).$ 
In addition, for $1 < \beta < 2$, we choose the exact solution to of \eqref{eq:multi_diffusion_wave_equations} with $m =1$ as 
$u(x,t)=(t^{\beta}+1) \sin(4\pi x)$,
with initial value condition $s(x)=\sin(4\pi x)$ 
and the source term $f( x,t ) =[ \Gamma ( \beta +1) +16\pi ^2( t^{\beta}+1 ) ] \sin(4\pi x).$

\begin{figure}[htb!]
	\centering
	\includegraphics[width=12.83cm,height=3.46cm]{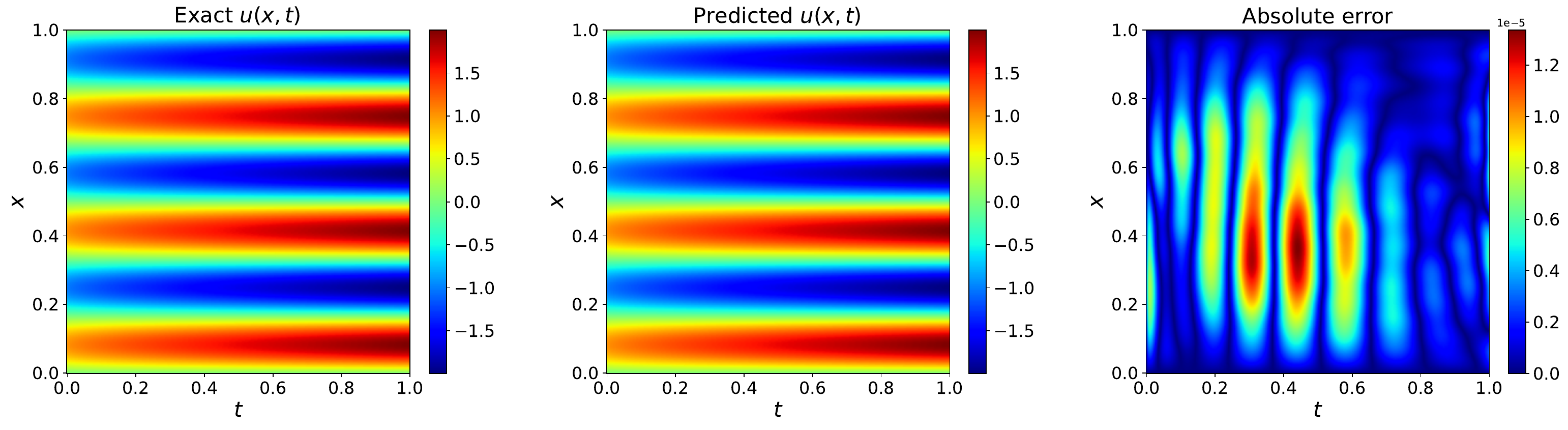}
	\includegraphics[width=12.8cm,height=3.46cm]{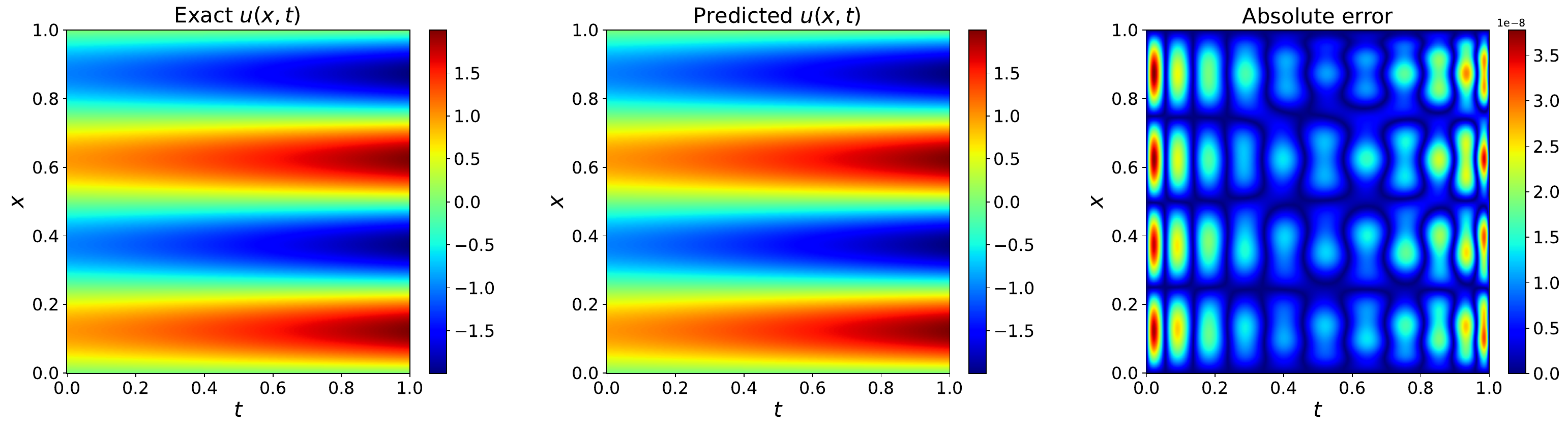}
	\caption{Numerical results of equation \eqref{eq:multi_diffusion_wave_equations} 
		with $m = 1$ and $u=(t^{\beta}+1) \sin(6\pi x)$ for $\mu=\beta = 0.70$ (top row)
		and $u=(t^{\beta}+1) \sin(4\pi x)$ for $\mu = \beta = 1.20$ (bottom row): 
		the exact solutions (left column), the predictions (middle column) 
		and the absolute errors of approximation (right column).}\label{ex1:fig_frequency_6}
\end{figure}

\begin{figure}[htb!]
	\centering
	\includegraphics[width=11.8cm,height=4.55cm]{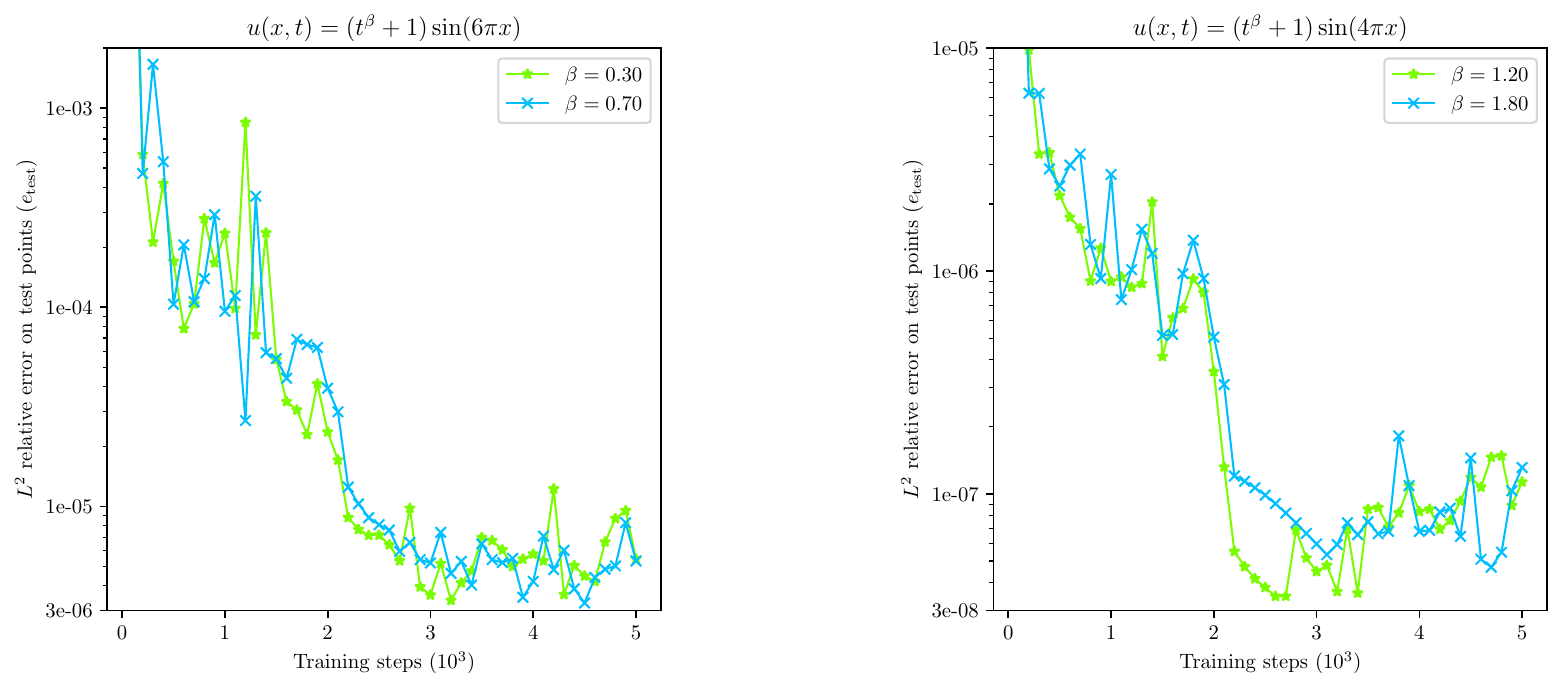}
	\caption{The relative $L^2$ error during the 
		training process for solving \eqref{eq:multi_diffusion_wave_equations}
		with $m =1$, $u=(t^{\beta}+1) \sin(6\pi x)$ and  $u=(t^{\beta}+1) \sin(4\pi x)$.}\label{fig:example1_high_frequency}
\end{figure}

Obviously, the solutions of this differential equation have relatively  
high oscillation.  We conduct the decomposition $u (x, t) = \widehat {u} (x, t) + s(x)$ 
and build the TNN function to approximate $\widehat {u} (x, t)$ with the 
homogeneous initial value condition. 
The corresponding numerical results are shown in  
Figures \ref{ex1:fig_frequency_6} and \ref{fig:example1_high_frequency}, 
Tables \ref{tab:frequency_6} and \ref{tab:frequency_4}, which indicate that 
the TNN subspace method proposed in this paper has high accuracy.
\begin{table}[htb]
	\centering
	\caption{Errors of equation \eqref{eq:multi_diffusion_wave_equations} 
		with $m = 1$, $\beta=0.7$ and $u=(t^{\beta}+1) \sin(6\pi x)$ using different methods.}\label{tab:frequency_6}
	\begin{tabular}{ccccc} 
		\toprule
		fPINNs  & IFPINNs & MC-fPINNs & Improved MC-fPINNs  &  TNN ($\mu = \beta $)   \\  
		\midrule
		5.0e-01 & 2.9e-02 & 6.4e-03 & 4.5e-03  &  5.2e-06    \\
		\bottomrule
	\end{tabular}
\end{table}
\begin{table}[htb]
	\centering
	\caption{Errors of equation 
		\eqref{eq:multi_diffusion_wave_equations} 
		with $m = 1$, $\beta=1.2$ and $u=(t^{\beta}+1) \sin(4\pi x)$ using different methods.}\label{tab:frequency_4}
	\begin{tabular}{ccc} 
		\toprule
		fPINNs  & IFPINNs  &  TNN ($\mu = \beta $) \\  
		\midrule
		6.6e-02  & 6.6e-02& 5.9e-08  \\
		\bottomrule
	\end{tabular}
\end{table}

In the numerical tests presented above, the proposed method demonstrates higher accuracy compared to MC-fPINNs and Improved MC-fPINNs. This improvement is primarily attributed to the fact that, when evaluating the loss function, the involved integrations can be reduced to low-dimensional ones due to the time-space separation of the time-dependent problem and the variable-separated structure of TNN. This reduction enables the use of classical quadrature schemes, which offer high integration accuracy. In addition, the alternating optimization strategy employed in the proposed algorithm further contributes to the overall accuracy.

On the other hand, although the MC integration method used in common PINN-related algorithms often suffers from higher variance than classical quadrature, potentially leading to reduced accuracy in loss function evaluation, it offers greater flexibility with respect to the neural network architecture. Owing to this property, both MC-fPINNs and Improved MC-fPINNs can accommodate arbitrary network structures, including TNN, thus providing broader applicability across a wide range of problems. In contrast,  the machine learning method proposed in this paper requires the special architecture of TNN.

\subsubsection{Multi-term time-fractional diffusion-wave equations} \label{example:multi-term FPDEs}
Here, let us consider solving the following one-dimensional multi-term time fractional 
diffusion equations \eqref{eq:multi_diffusion_wave_equations}.
For this aim, we choose the exact solution to \eqref{eq:multi_diffusion_wave_equations} with $m=2$ 
as $u(x,t)=(t^{\alpha_2}+t^{\alpha_1}) \sin(2\pi x)$, $\alpha _1\leqslant \alpha _2$,  
$\alpha_1, \alpha_2 \in (0,4)$, $\beta_2 < \alpha_{1} + 0.5$ by setting the initial value condition $s(x)=0$ 
and the source term
\begin{eqnarray}\label{multi_source_term}
	f(x,t)=\left[ \sum_{i=1}^2{\sum_{j=1}^2{\frac{\Gamma (1+\alpha_j)
				t^{\alpha _j-\beta _i}}{\Gamma (1+\alpha _j-\beta _i)}}}+4\pi^2(t^{\alpha _2}
	+t^{\alpha _1}) \right] \sin(2\pi x). 
\end{eqnarray}

\begin{table}[htb]
	\centering
	\caption{Errors of equation \eqref{eq:multi_diffusion_wave_equations} with $m=2$ for different values of $\beta_{1},\beta_{2},\alpha_1,\alpha_2$  using TNN ($\mu =\beta_{1}$).}\label{tab:ex.two_term}  
	\begin{tabular}{cccccc} 
		\toprule
		\multicolumn{3}{c}{$\beta_1,\beta_2 \in (0,1)$} & \multicolumn{3}{c}{$\beta_1,\beta_2 \in (1,2)$} \\ 
		\cmidrule(lr){1-3} \cmidrule(lr){4-6}
		$(\beta_{1},\beta_{2}) $  & $(\alpha_1,\alpha_2)$ &  $e_{\rm test}$ & $(\beta_{1},\beta_{2}) $  & $(\alpha_1,\alpha_2)$ & $e_{\rm test}$  \\ 
		\midrule
		(0.20 , 0.30)  & (0.40 , 0.80) &  1.747e-05 & (1.10 , 1.65)  & (1.20 , 1.80)   &  5.009e-05\\
		(0.10 , 0.60)  & (0.70 , 0.90) &  2.896e-06 &  (1.40 , 1.50)  & (1.20 , 1.80)    &  3.509e-05\\
		\cdashline{1-6}[1pt/1pt] 
		(0.20 , 0.50)  & (0.10 , 1.00) &  1.907e-05 &  (1.10 , 1.70)  & (2.20 , 2.40)  &  5.314e-07\\
		(0.60 , 0.80)  & (0.40 , 0.80) &  8.647e-05 &  (1.30 , 1.90)  & (2.20 , 2.40)  &  9.042e-07\\
		\cdashline{1-6}[1pt/1pt]  
		(0.50 , 0.80)  & (1.20 , 1.40) &  1.190e-06 &  (1.10 , 1.50)  & (3.50 , 3.60)   &  1.193e-07\\ 	
		(0.10 , 0.50)  & (2.20 , 2.40) &  2.692e-07 &  (1.30 , 1.80)  & (3.50 , 3.60)  &  2.686e-08\\	 	 		 
		\bottomrule
	\end{tabular}
\end{table}

The corresponding numerical results are shown in Table \ref{tab:ex.two_term}. 
From this table, it can be seen that the relative error attains around $1 \times 10^{-5}$ for $\beta_{1},\beta_{2}  \in (0,1)$ when $\alpha_ {1}, \alpha_{2} \in (0, 1)$ or $\beta_{1},\beta_{2}  \in (1,2)$ when $\alpha_ {1}, \alpha_{2} \in (1, 2)$ .  And the relative error increase with the improvement of the smoothness of the exact solution.

\begin{table}[htb!]
	\centering
	\caption{Errors of equation \eqref{eq:multi_diffusion_wave_equations} with $m=4$ for different values of 
		$\beta_{1},\beta_{2} \in (0,1),\beta_{3},\beta_{4} \in (1,2),\alpha_1, \alpha_2 \in (0,4)$ using TNN ($\mu =\beta_{3}$).}\label{tab:ex.1_Multiterm}  
	\begin{tabular}{ccc} 
		\hline
		$\boldsymbol{\beta} =(\beta_{1},\beta_{2},\beta_{3},\beta_{4}) $  & $(\alpha_1,\alpha_2)$  & $e_{\rm test}$   \\ \cline{1-3} 
		(0.20 , 0.70 , 1.10 , 1.20)  & (0.95 , 1.90)  & 1.422e-05 \\
		(0.20 , 0.30 , 1.10 , 1.45)  & (1.30 , 1.90)  & 9.082e-06 \\	
		\cdashline{1-3}[1pt/1pt] 
		(0.30 , 0.60 , 1.30 , 1.45)  & (2.10 , 2.50)  & 1.243e-07\\
		(0.40 , 0.90 , 1.40 , 1.85)  & (3.20 , 3.60)  & 9.587e-08 \\
		\hline
	\end{tabular}
\end{table}
\begin{figure}[htb!]
	\centering
	\includegraphics[width=11.8cm,height=4.55cm]{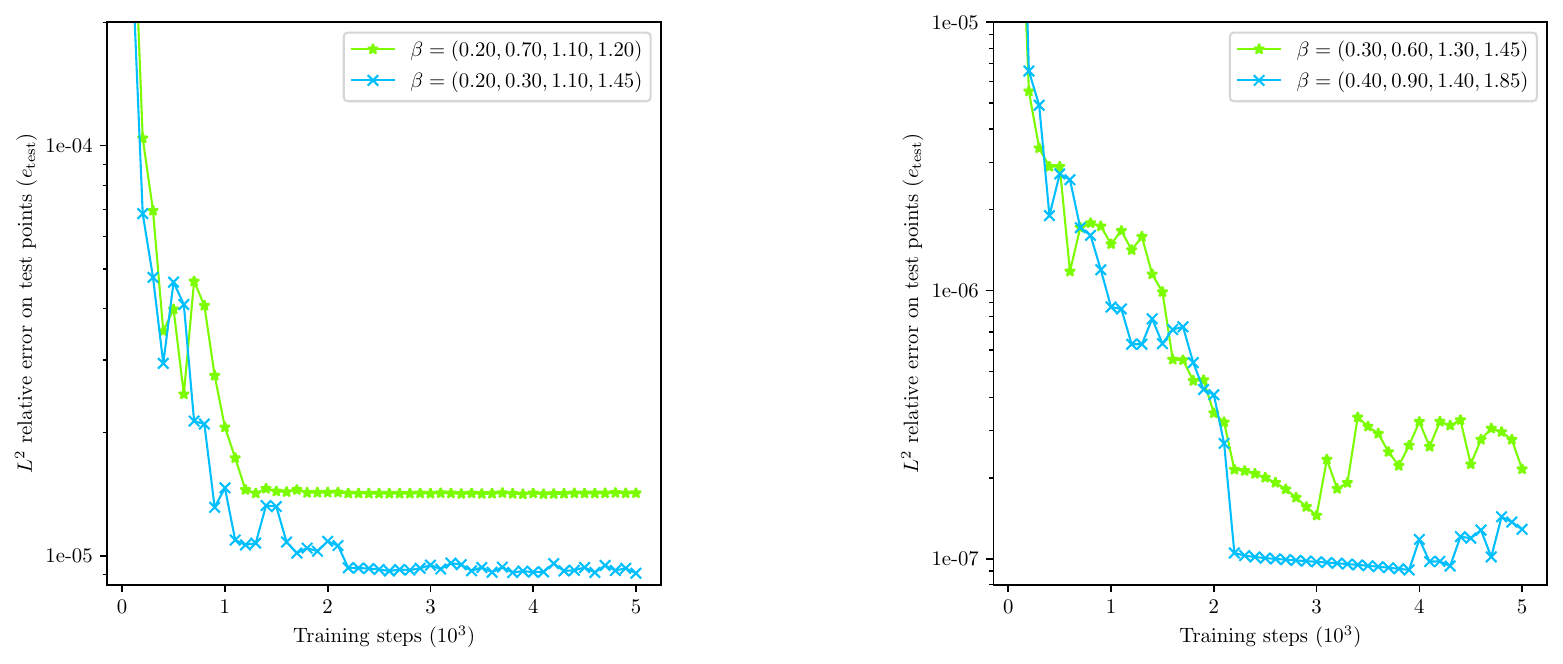}
	\caption{The relative $L^2$ errors during the
		training process for solving \eqref{eq:multi_diffusion_wave_equations} with $m = 4$, $\beta_{1},\beta_{2} \in(0,1)$, $\beta_{3}, \beta_{4}\in(1,2)$ and  $u=(t^{\alpha_{1}}+t^{\alpha_{2}}) \sin(2\pi x)$}\label{fig:example1_multi}
\end{figure}
The next example is to consider solving \eqref{eq:multi_diffusion_wave_equations} with $m=4$.  
We set the initial value condition $s(x)=0$ and the source term 
\begin{eqnarray}\label{multi_forth_source_term}
	f(x,t)=\left[ \sum_{i=1}^4{\sum_{j=1}^2{\frac{\Gamma (1+\alpha _j)t^{\alpha _j-\beta _i}}
			{\Gamma (1+\alpha _j-\beta _i)}}}+4\pi^2(t^{\alpha _2}+t^{\alpha _1}) \right] \sin(2\pi x), 
\end{eqnarray}
such that the exact solution is $u(x,t)=(t^{\alpha_2}+t^{\alpha_1}) \sin(2\pi x)$ 
for $\alpha_1\leqslant \alpha_2$,  $\alpha_1, \alpha_2 \in (0,1)$ and $\beta_{4} < \alpha_{1} + 0.5$.

For $\beta_{1}, \beta_{2}\in (0,1)$, $\beta_{3}, \beta_{4} \in (1,2)$, the corresponding numerical 
results are shown in Table \ref{tab:ex.1_Multiterm} and 
Figure \ref{fig:example1_multi}, which indicate that the high efficiency  
of the TNN method. 

\subsection{Nonlinear TFPIDEs with Fredholm integral}\label{section:nonlinear}
In this subsection, we focus on solving one-dimensional single/multi-term nonlinear TFPIDEs (\ref{FPIDE_problem}) with Fredholm integral. 

Due to space constraints, this subsection presents two one-dimensional nonlinear examples and higher-dimensional cases are discussed in  Subsection \ref{section:nonlinear_Volterra}.

\subsubsection{Single-term time-fractional Fredholm IDE} \label{example3}
Here, let us consider the following single-term TFPIDE, which includes a nonlinearity embedded within the Fredholm integral. This setup is designed to validate the capability of the proposed method in solving TFPIDEs involving nonlinear integral terms: Find $u(x,t)$ such that 
\begin{equation}\label{eq:single_1d_Fredholm}
	\left\{ \begin{aligned}
		_{0}^{C}D_{t}^{\beta}u(x,t)&=\frac{\partial ^2u(x,t)}{\partial x^2}+f(x,t)\\
		&+\frac{1}{2}\cos(x)\int_{-\frac{\pi}{2}}^{\frac{\pi}{2}}{stu^2(s,t)}ds, \ \ (x,t)\in [-\pi /2,\pi /2]\times (0,1],\\
		u(x,0)&=0, \ \ x\in [-\pi /2,\pi /2],\\
		u(-\pi /2,t)&=u(\pi /2,t)=0, \ \ t\in (0,1],\\
	\end{aligned} \right. 
\end{equation}
where $\beta\in(0,1)$. 

For the first example, the exact solution is set to be $u(x,t)= (t^{\beta}+t)\cos (x)$ by choosing the source term 
$$f(x,t)=\left[ \Gamma ( 1+\beta ) +1 / \Gamma (2-\beta ) t^{1-\beta}
+( t^{\beta}+t) \right] \cos(x).$$
Also, the exact solution for the second example is $ u(x,t)=t^\beta\cos(x)$ \cite{zhang2024adaptive} 
by setting the source term as 
$$f(x,t)=\Gamma (1+\beta)\cos(x)+t^{\beta}\cos(x).$$

\begin{figure}[htb!]
	\centering
	\includegraphics[width=11.8cm,height=4.55cm]{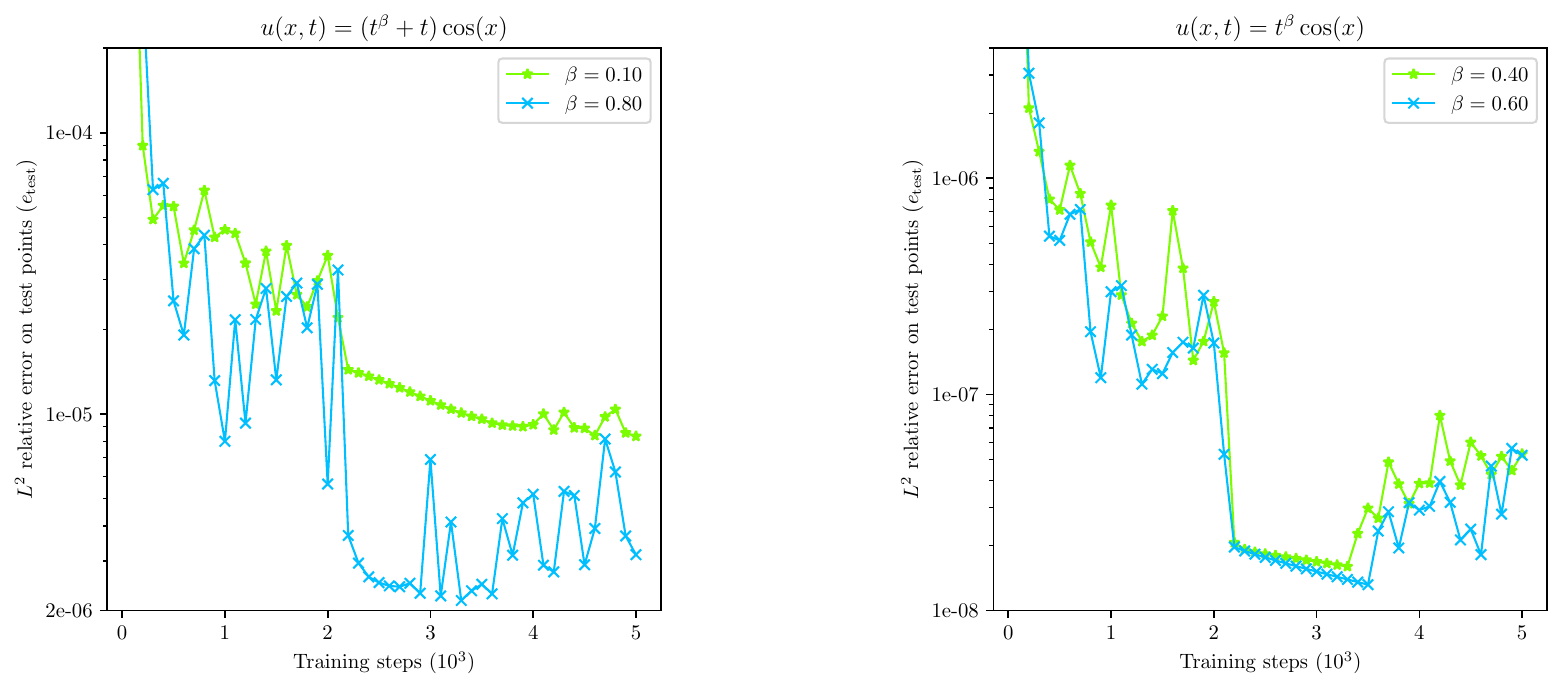}
	\caption{The relative $L^2$ error  during the
		training process for solving \eqref{eq:single_1d_Fredholm}.}\label{fig:single_1d_Fredholm_error} 
\end{figure}
\begin{figure}[htb!]
	\centering
	\includegraphics[width=12.8cm,height=3.46cm]{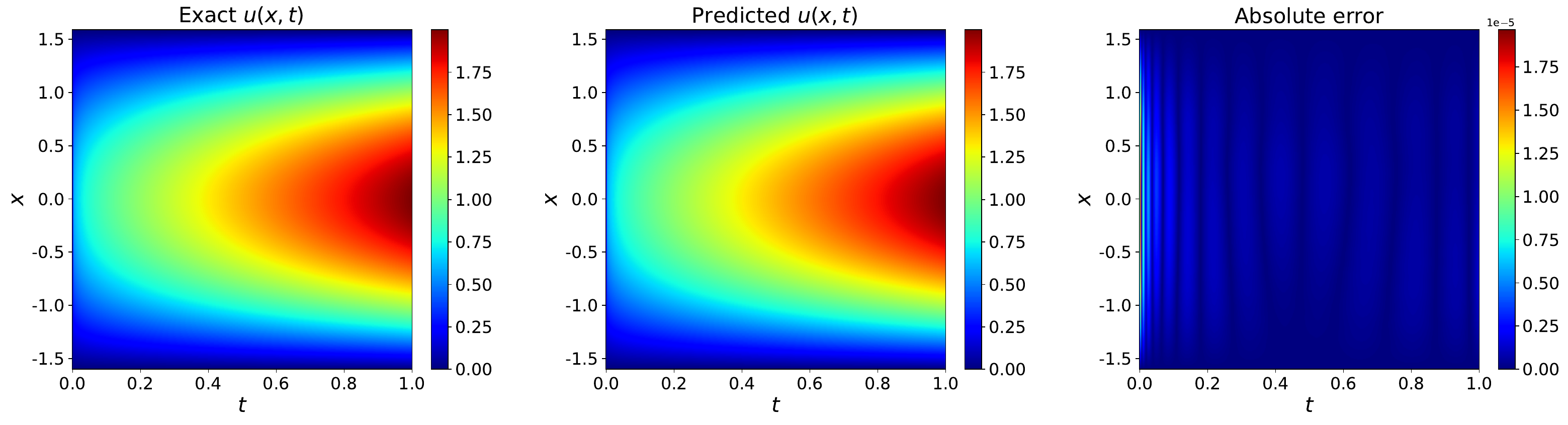}
	\includegraphics[width=12.8cm,height=3.46cm]{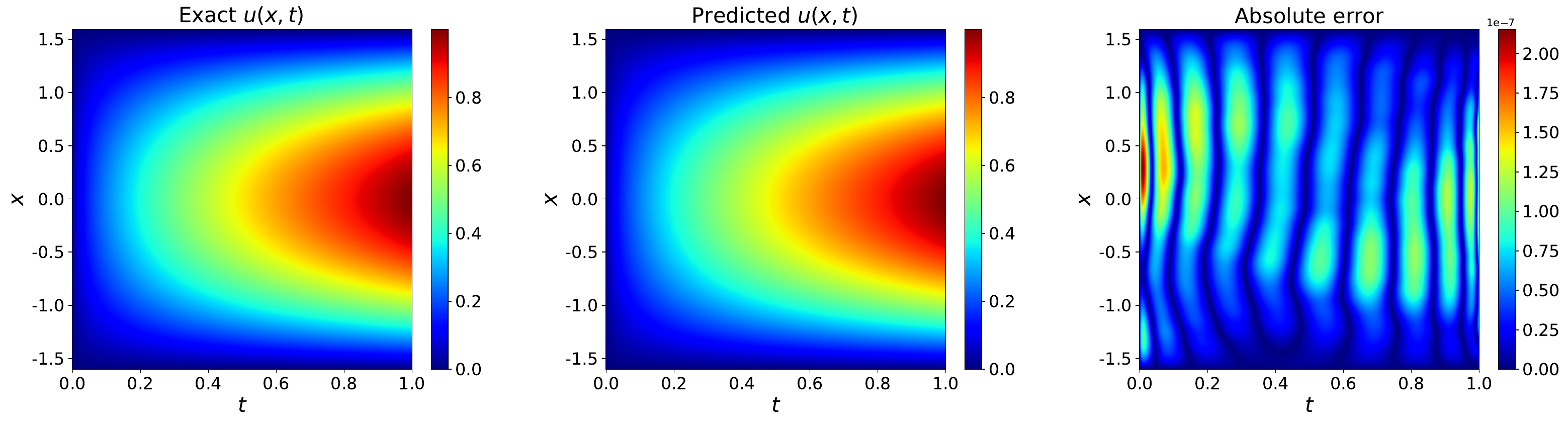}
	\caption{Numerical results of equation \eqref{eq:single_1d_Fredholm} for 
		$u=(t^{\beta}+t )\cos(x)$, $\beta = 0.10$ (top row) and $u= t^{\beta}\cos(x)$, $\beta = 0.60$ (bottom row): 
		the exact solutions (left column), 
		the predictions (middle column), and the absolute errors (right column).}\label{fig:single_1d_Fredholm} 
\end{figure}

The following formula gives the numerical scheme for the Fredholm integral in \eqref{eq:single_1d_Fredholm}
\begin{eqnarray*}
	\int_{-\frac{\pi}{2}}^{\frac{\pi}{2}}{\cos(x)ts\Psi ^2(s,t)}ds
	\approx \sum_{i=1}^p{\sum_{j=1}^p{c_i c_j\cos(x)
			\sum_{k=1}^{N_s}{w_ks_k\phi_{1,i}(s_k)\phi_{1,j}(s_k)t^{2\mu +1}\phi_{t,i}(t)\phi_{t,j}(t)}}}, 
\end{eqnarray*}
where $\{w_k,s_k\}_{k=1}^{N_m}$ are the nodes and weights of Gauss-Legendre quadrature 
in the interval $[-\pi/2,\pi/2]$, which is subdivided into 25 subintervals, with 16 Gauss points selected within each subinterval for high accuracy.
\begin{table}[htb]
	\centering
	\caption{Errors of equation (\ref{eq:single_1d_Fredholm}) for different $\beta$ values using different methods.} 
	\label{tab:single_1d_Fredholm} 
	\begin{tabular}{cccccc} 
		\toprule
		\multicolumn{4}{c}{$u = t^\beta \cos(x)$} & \multicolumn{2}{c}{$u = (t^\beta + t)\cos(x)$} \\ 
		\cmidrule(lr){1-4} \cmidrule(lr){5-6}
		$\beta$ & AO-fPINNs  & AWAO-fPINNs  & TNN & $\beta$ & TNN \\  
		\midrule
		0.4    & 1.001e-3   &  6.825e-4   &  8.063e-08  &  0.1    & 8.688e-06  \\
		0.6    & 1.047e-3   &  5.550e-4   &  4.415e-08  & 0.8     & 3.348e-06 \\
		\bottomrule
	\end{tabular}
\end{table}

The corresponding numerical results are presented in Figures \ref{fig:single_1d_Fredholm_error} and \ref{fig:single_1d_Fredholm}, and Table \ref{tab:single_1d_Fredholm},
which show the relative $L^2$ error during the
training process and curves of the numerical results for the TNN method.  Table \ref{tab:single_1d_Fredholm} shows that the relative error can reach $1 \times 10^{-8}$ for the case of the exact solution $u(x, t) = t^\beta\cos(x)$. 
The numerical comparison  is included in Table \ref{tab:single_1d_Fredholm}, which shows that AO-fPINNs and AWAO-fPINNs \cite{zhang2024adaptive} can only obtain the accuracy of $1 \times 10^{-3}$ or $1 \times 10^{-4}$, while the relative error of the TNN method can also arrive at the accuracy of $1 \times 10^{-8}$.

\subsubsection{Two-term TFPDE with weakly singular Fredholm integral} \label{example:singular_Fredholm}
To demonstrate the adaptability of our method for solving TFPIDEs with weakly singular Fredholm integrals, we consider the following two-term TFPIDE for $\beta_{1}$, $\beta_{2}\in (0,1)$: 
Find $u(x,t)$ such that 
\begin{equation}\label{eq:singular_Fredholm}
	\left\{ \begin{aligned}
		\sum_{k=1}^2{_{0}^{C}\mathrm{D}_{t}^{\beta_{k}}u(x,t)}&=\frac{\partial^2u(x,t)}{\partial x^2}
		+f(x,t,u)\\
		&+\int_0^1{|}t-s|^{-1/2}u(x,s)ds, \ \ (x,t)\in (0,1)\times (0,1],\\
		u(x,0)&=0,\ \ x\in [0,1],\\
		u(0,t)&=u(1,t)=0,\ \ t\in (0,1].
	\end{aligned} \right. 
\end{equation}

For the weakly singular Fredholm integral in this example, we decompose it into two Volterra integrals as follows
\begin{eqnarray*}
	\int_0^1{|}t-s|^{-1/2}s^{\mu}\phi _{1,j}(s)ds=\int_t^1{(s-t)^{-1/2}}s ^{\mu}
	\phi_{1,j}\left( s \right) ds+\int_0^t{(t-s)^{-1/2}}s^{\mu}\phi_{1,j}(s)ds.
\end{eqnarray*}
And this integral can be discreted after performing the transformations of $s=t+(1-t) m$ and $s=tm$ as follows
\begin{eqnarray*}
	&&\int_0^1{|}t-s|^{-1/2}s^{\mu}\phi _{t,j}(s)ds\\
	&=&(1-t)^{1/2}\sum_{k=1}^{N_m}{w_{k}^{(0,-1/2 )}}[t+(1-t)m_{k}^{(0,-1/2 )}]^{\mu}
	\phi_{t,j}(t+(1-t)m_{k}^{(0,-1/2 )})\\
	&&+t^{1/2 +\mu}\sum_{k=1}^{N_m}{w_{k}^{(-1/2 ,\mu )}\phi _{t,j}(tm_{k}^{(-1/2 ,2\mu )})}.
\end{eqnarray*}
Then the Fredholm integral in this example can be calculated as follows: 
$$\int_0^1{|}t-s|^{-1/2}\Psi (x,s)ds=\sum_{j=1}^p{c_j}\phi_{1,j}(x)
\int_0^1{|}t-s|^{-1/2}s^{\mu}\phi _{t,j}(s)ds.$$

\begin{figure}[htb!]
	\centering
	\includegraphics[width=11.8cm,height=4.55cm]{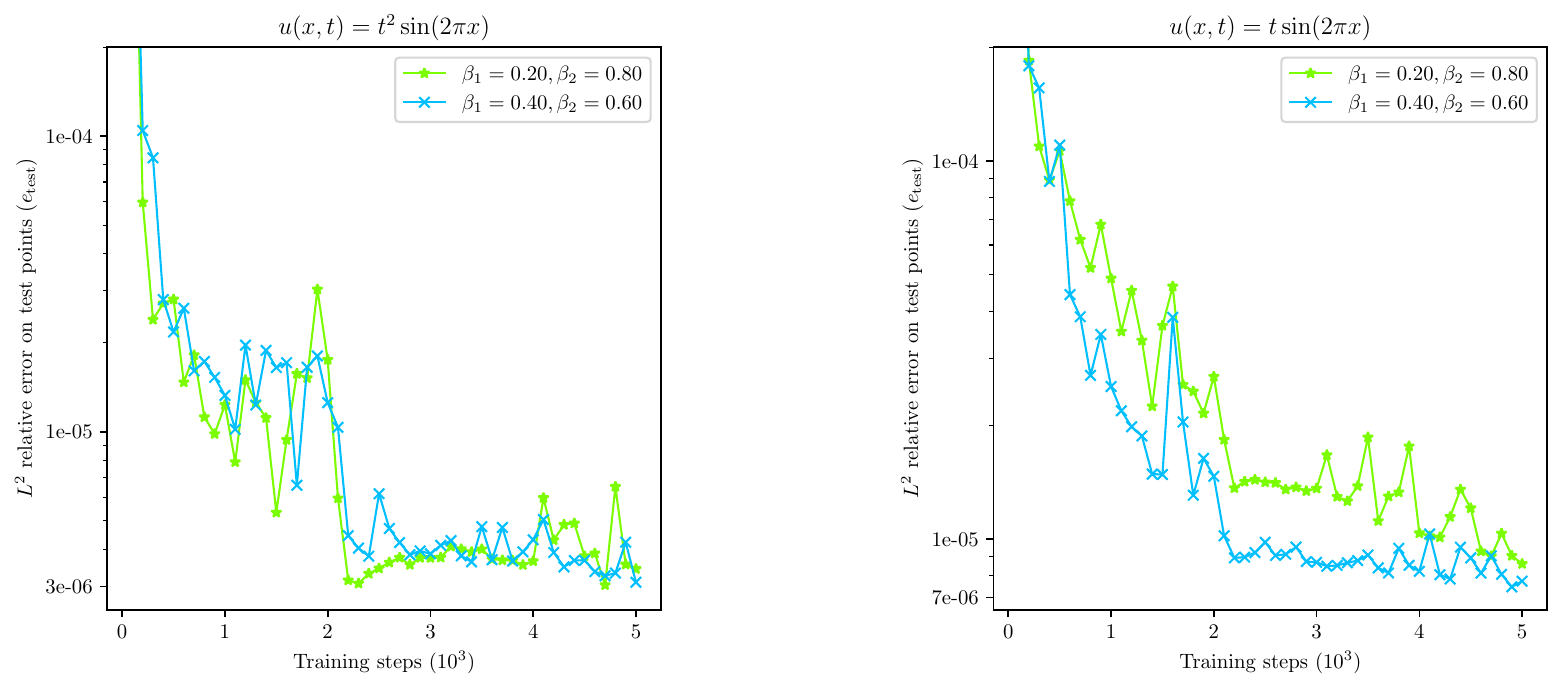}
	\caption{The relative $L^2$ error  during the
		training process for solving \eqref{eq:singular_Fredholm}.}\label{fig:singular_Fredholm} 
\end{figure}

\begin{figure}[htb!]
	\centering
	\includegraphics[width=12.8cm,height=3.46cm]{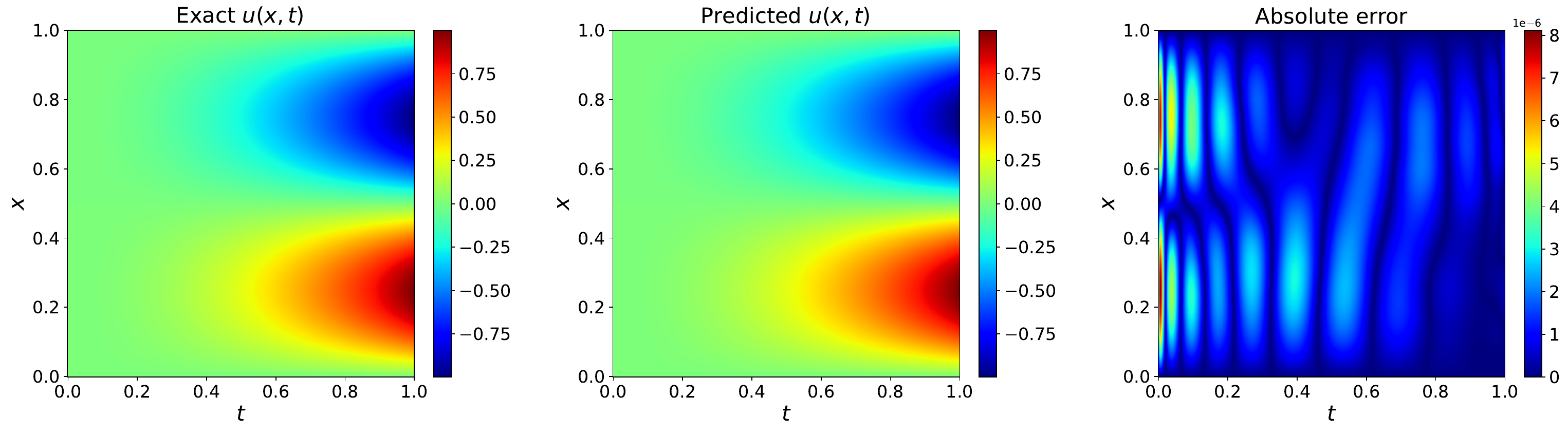}
	\includegraphics[width=12.8cm,height=3.46cm]{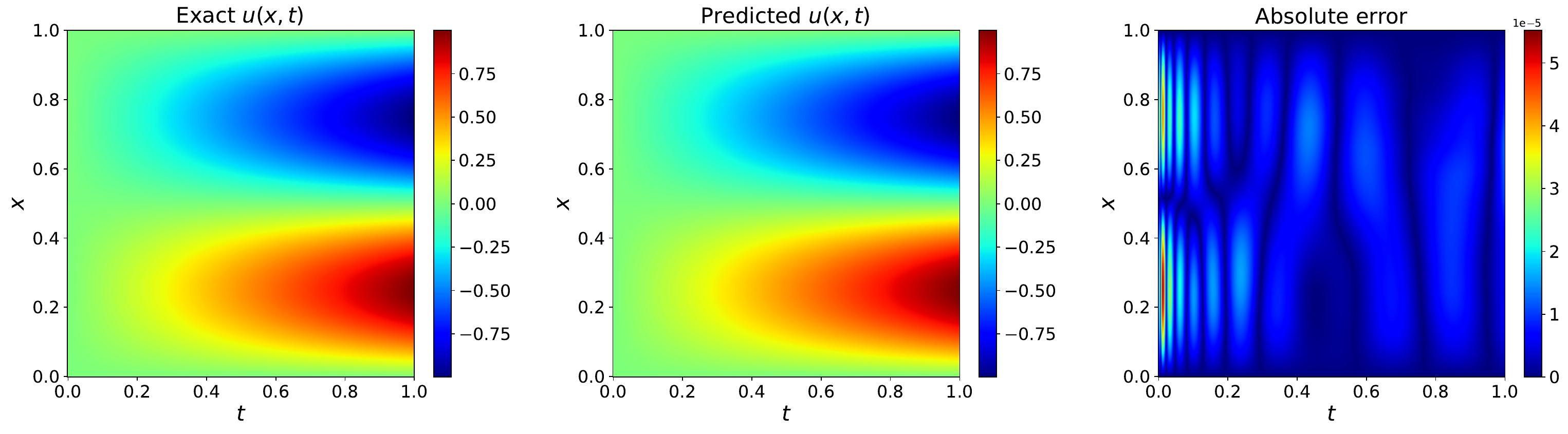}	
	\caption{Numerical results of equation \eqref{eq:singular_Fredholm} with  
		$u=t^{2} \sin(2\pi x)$ for $\beta_{1} = 0.20$, $\beta_{2} = 0.80$ (top row) 
		and $u=t \sin(2\pi x)$ for $\beta_{1} = 0.20$, $\beta_{2} = 0.80$ (bottom row): 
		the exact solutions (left column), the predictions (middle column), and the absolute errors 
		of approximation (right column).}\label{fig:ex.add} 
\end{figure}

Set the exact solution to \eqref{eq:singular_Fredholm} as $ u(x, t) = t^{2} \sin(2\pi x) $ and the source term is chosen as
\begin{eqnarray*}
	f(x,t,u)&=&\sum_{t=1}^2{\frac{\Gamma (3)}{\Gamma (3-\beta _i)}t^{2-\beta _i}}
	\sin\mathrm{(}2\pi x)+4\pi ^2t^2\sin(2\pi x)\\
	&&-\left[  \frac{16}{15}t^{5/2}+2(1-t)^{1/2}t^2+\frac{4}{3}(1-t)^{3/2}t+\frac{2}{5}(1-t)^{5/2}\right] \sin(2\pi x).
\end{eqnarray*}
In addition, set the exact solution to \eqref{eq:singular_Fredholm} as $ u(x,t) = t \sin(2\pi x)$ and the source term is given by  
$$f(x,t,u)=\left[ \sum_{i=1}^2{\frac{\Gamma (2)}{\Gamma (2-\beta _i)}t^{1-\beta _i}}+4\pi ^2t-\frac{4}{3}t^{\frac{3}{2}}-2t(1-t)^{\frac{1}{2}}-\frac{2}{3}(1-t)^{\frac{3}{2}} \right] \sin(2\pi x).$$

\begin{table}[htb]
	\centering
	\caption{Errors of equation \eqref{eq:singular_Fredholm} for $\beta_1,\beta_2 \in (0,1)$ using TNN ($\mu = \beta_{1}$).}\label{tab:singular_Fredholm}  
	\begin{tabular}{ccc} 
		\toprule
		$(\beta_1,\beta_2)$   & $u=t^{2}\sin ( 2\pi x )$   & $u =t\sin ( 2\pi x )$ \\  
		\midrule
		(0.2,0.8)   &  4.801e-06    & 8.547e-06 \\
		(0.4,0.6)   &  3.290e-06    & 7.555e-06 \\
		\bottomrule
	\end{tabular}
\end{table}

The corresponding numerical results are shown in Figures \ref{fig:singular_Fredholm} and \ref{fig:ex.add}, 
Table \ref{tab:singular_Fredholm}, which show that the relative error can reach the high accuracy of \(1 \times 10^{-6}\).
\subsection{Nonlinear TFPIDEs with Volterra integral}\label{section:nonlinear_Volterra}
In this subsection, we focus on solving one and three-dimensional single/multi-term nonlinear TFPIDEs (\ref{FPIDE_problem}) involving Volterra integrals.

\subsubsection{Single-term TFPIDE with double Volterra integrals}\label{single_term_1d_volterra}
We consider the following nonlinear TFPIDE with non-homogeneous boundary conditions to validate the robustness of our method in solving TFPIDEs with non-homogeneous boundary conditions:
Find $u(x,t)$ such that 
\begin{equation}\label{eq:single_term_1d_volterra}
	\left\{ \begin{aligned}
		_{0}^{C}D_{t}^{\beta}u(x,t)=&\frac{\partial ^2u(x,t)}{\partial x^2}+f(x,t,u)\\
		&+\int_0^t{\int_0^x{\tau}}e^{x-s}u(s,\tau )dsd\tau,\ \ (x,t)\in (0,1)\times [0, 1],\\
		u(x,0)=&e^x,\ \ x\in [0,1],\\
		u(0,t)=&(t-1)^2, u(1,t)=e(t-1)^2,\ \ t\in (0,1], 
	\end{aligned} \right. 
\end{equation}
where $\beta \in (0,1)$. The source term is set to be 
\begin{eqnarray*}
	f(x,t,u)&=&e^x\left( \frac{2}{\Gamma (3-\beta )}t^{2-\beta}-\frac{2}{\Gamma (2-\beta )}t^{1-\beta} \right) 
	-xe^x\left( \frac{1}{4}t^4-\frac{2}{3}t^3+\frac{1}{2}t^2 \right)\\
	&&+e^x(t-1)^2(e^x(t-1)^2-1), 
\end{eqnarray*}
such that the exact solution to \eqref{eq:single_term_1d_volterra} 
is $u(x,t)=e^{x}\left(t-1\right)^{2}$. 

By using the mappings $\tau = xn$ and $s = tm$, the interval of the Volterra integral is transformed to 
$[0,1]$ and then the Gauss-Jacobi quadrature scheme is used for computing 
double integral in \eqref{eq:single_term_1d_volterra} of the corresponding TNN approximation $\Psi(x,t,c,\theta)$ as follows
$$\int_0^t{\int_0^x{\tau}e^{x-s}\Psi dsd\tau}\approx \sum_{j=1}^p{c_j\sum_{i=1}^{N_n}{w_{i}^{n}e^{x(1-n_i)}x\phi _{1,j}(xn_i)\sum_{k=1}^{N_m}{w_{k}^{m}m_{k}^{1+\mu}t^{2+\mu}\phi _{t,j}(tm_k)}},}$$
where $\{w_{i}^{n}, n_i\}_{i=1}^{N_m}$ and $\{w_{k}^{m},m_k\}_{k=1}^{N_m}$ 
denote the weights and nodes of Gauss-Legendre quadrature scheme on the 
interval $[0,1]$. This interval is subdivided into 25 subintervals, with 16 Gauss points selected within each subinterval to ensure sufficient accuracy for discretizing the Volterra integral.
\begin{figure}[ht!]
	\centering
	\includegraphics[width=12.8cm,height=3.46cm]{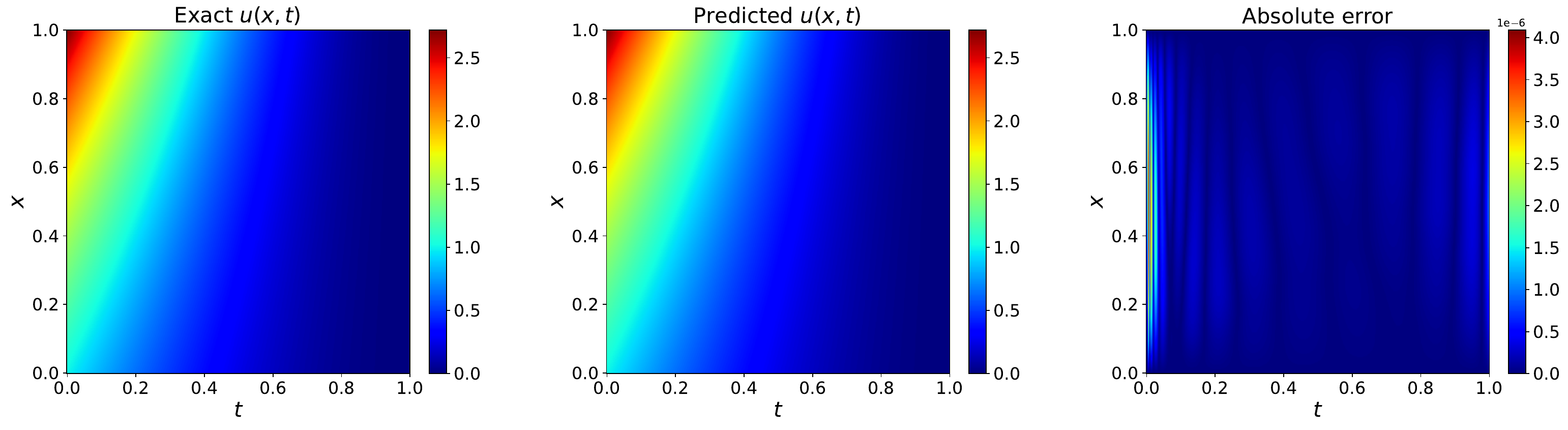}
	\caption{Numerical results of equation \eqref{eq:single_term_1d_volterra} for $\beta = 0.20$: the exact solutions (left column), 
		the predictions (middle column), and the absolute errors of approximation (right column).}\label{Figure:single_term_1d_volterra} 
\end{figure}
\begin{figure}[ht!]
	\centering
	\includegraphics[width=11.8cm,height=4.55cm]{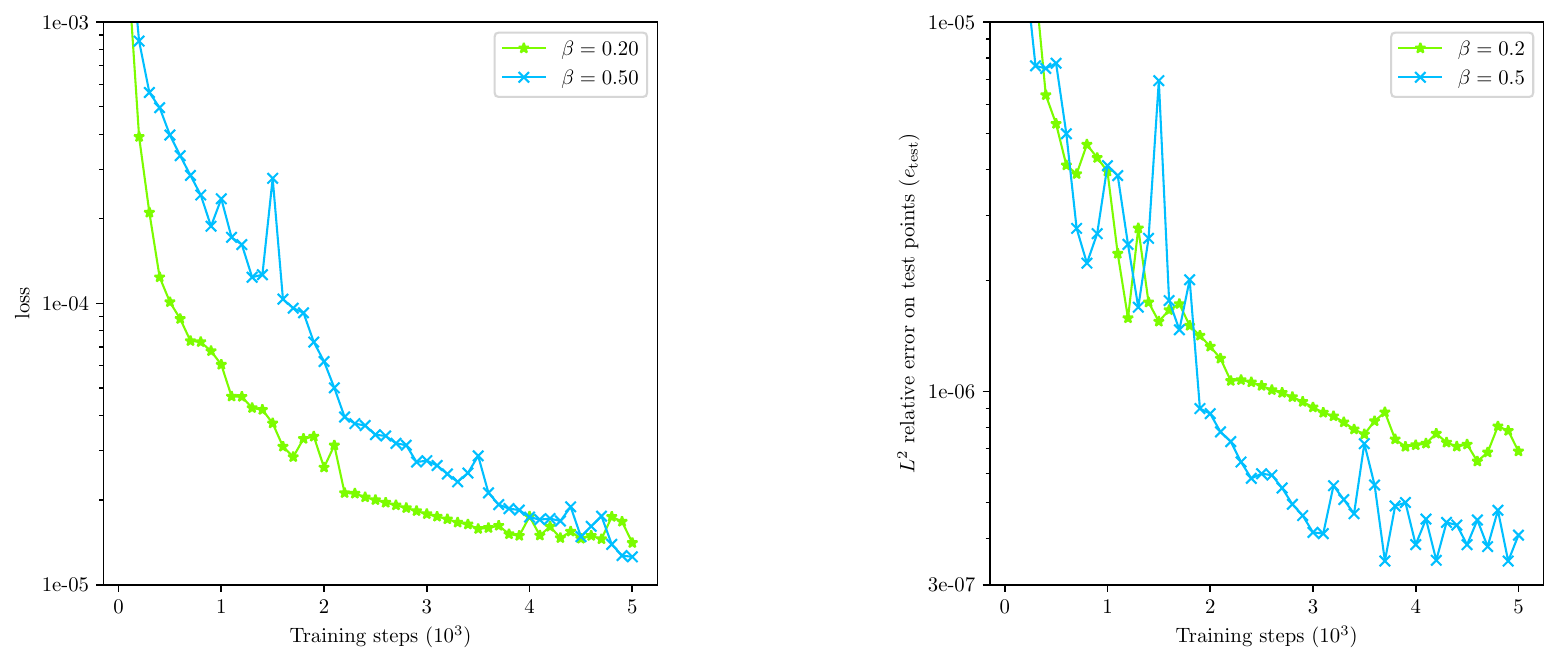}
	\caption{The loss curve and relative $L^2$ error  during the
		training process for solving  \eqref{eq:single_term_1d_volterra}.}\label{fig:single_term_1d_volterra}
\end{figure}

We construct a new function $\widehat{g}(x, t) = [(e-1)x+1]\left( t-1 \right) ^2$ 
and equation \eqref{eq:single_term_1d_volterra} can be transformed 
into the one for the exact solution $\widehat{u}(x,t) $ with the homogeneous initial boundary
value conditions by the following decomposition
$$ u( x,t ) =\widehat{u}(x,t) +e^x-(e-1)x-1+\widehat{g}(x,t).$$ 

\begin{table}[htb!]
	\centering
	\caption{Errors of equation \eqref{eq:single_term_1d_volterra} for different values of  $\beta $ using different methods .}\label{tab:ex_single_term_1d_volterra} 
	\begin{tabular}{cccc} 
		\hline
		$\beta$ & AO-fPINNs   & AWAO-fPINNs  &   TNN ($\mu =\beta$)  \\ \cline{1-4} 
		$0.2$    & 1.503e-3   &  5.599e-4   & 6.582e-07   \\
		$0.5$    & 9.265e-4   &  2.455e-4   & 3.486e-07  \\
		\hline
	\end{tabular}
\end{table} 

During the training process, the corresponding numerical results are presented 
in Figures~\ref{Figure:single_term_1d_volterra} and \ref{fig:single_term_1d_volterra}, Table~\ref{tab:ex_single_term_1d_volterra}.  
These results demonstrate that the relative error achieve 
an high accuracy of order $10^{-7}$ with $\mu = \beta$, which is significantly 
superior to the accuracy achieved by AO-fPINNs and AWAO-fPINNs \cite{zhang2024adaptive}.

\subsubsection{Three-dimensional time-fractional IDE with quadruple Volterra integrals} \label{example:3d_Volterra}
Here, we consider the following three-dimensional time-fractional IDE with quadruple integrals and fractional order $\beta_{1},\beta_{2}\in(1,2)$: Find $u(\boldsymbol{x},t)$ such that
\begin{equation}\label{eq:3d_Volterra}
	\left\{ \begin{aligned}
		\sum_{k=1}^2{_{0}^{C}}D_{t}^{\beta _k}u(\boldsymbol{x},t)&=\Delta u(\boldsymbol{x},t)+f(\boldsymbol{x},t,u)
		+v(\boldsymbol{x},t,u),  (\boldsymbol{x},t)\in \Omega \times (0,T],\\
		v\left( \boldsymbol{x},t,u \right) &=\int_0^t{\int_0^{x_3}{\int_0^{x_2}{\int_0^{x_1}{\tau}}}(p-q)u(p,q,s,\tau )dpdqdsd\tau},\\
		u(\boldsymbol{x},t)&=0,  \boldsymbol{x}\in \partial \Omega ,  t\in (0,T],\\
		u(\boldsymbol{x},0)&=0,  \boldsymbol{x}\in \bar{\Omega}.\\
	\end{aligned} \right. 
\end{equation}
For the first example, we consider the domain $\Omega=(0,\pi)^{3}$, the time-interval $(0,1]$, and the exact solution  to \eqref{eq:3d_Volterra} as $ u(\boldsymbol{x},t)=t^{2+\beta_2}\prod_{i=1}^3{\sin(\pi x_i})$  
by setting the source term as 
\begin{eqnarray*}
	f(\boldsymbol{x},t,u)&=&\prod_{i=1}^3{\sin(\pi x_i})\left[ \sum_{k=1}^2{\frac{\Gamma (3+\beta _2)}{\Gamma (3+\beta _2-\beta _k)}t^{2+\beta _2-\beta _k}} \right]
	+\frac{t^{4+\beta _2}}{4+\beta _2}( \cos(x_3)-1 ) I(x_1,x_2),
\end{eqnarray*}
where  $I(x,y): =\sin(y-x) +\sin (x) -\sin(y) +(x-y) \cos(x) \cos(y) +y\cos(y) -x\cos(x)$.

For the second example, we consider the domain $\Omega=(0,1)^{3}$, the time-interval $(0,1]$, and choose the non-variable-separated exact solution to (\ref{eq:3d_Volterra}) as $u(\boldsymbol{x},t)=t^{2\beta_2}e^{x_1x_2x_3}\prod_{i=1}^3{\sin(\pi x_i})$. The source term  can be derived from the exact solution as 
\begin{eqnarray*}
	f(\boldsymbol{x},t,u)
	&=&e^{x_1x_2x_3}\prod_{i=1}^3{\sin(\pi x_i)\left[ \frac{\Gamma (2\beta _2+1)}{\Gamma (2\beta _2+1-\beta _1)}t^{2\beta _2-\beta _1}+\frac{\Gamma (2\beta _2+1)}{\Gamma (\beta _2+1)}t^{\beta _2} \right]}\\
	&&-e^{x_1x_2x_3}\prod_{i=1}^3{\sin(\pi x_i)\left[ (x_2x_3)^2+(x_1x_3)^2+(x_1x_2)^2-3\pi ^2 \right] t^{2\beta _2}}\\
	&&-2\pi t^{2\beta _2}e^{x_1x_2x_3}\left[ \left( J(x_2,x_3,x_1)+J(x_1,x_3,x_2)+J(x_1,x_2,x_3) \right) \right]\\
	&&- \frac{t^{2\beta _2+2}x_1x_2x_3}{2(\beta _2+1)}\int_{[0,1]^3}{(x_1y_1}-x_2y_2)e^{\prod_{i=1}^3{x_iy_i}}\prod_{i=1}^3{\sin(\pi x_iy_i)}\,d^3\boldsymbol{y},
\end{eqnarray*} 
where  $J(x,y,z):=xy\sin(\pi x)\sin(\pi y)\cos(\pi z)$.

The domain of the quadruple Volterra integrals can be transformed into the unit hypercube $[0,1]^{4}$ by applying the linear mappings $p = x_1m$, $q = x_2n$, $s = x_3h$, and $\tau = t \eta $. 
Then the Gauss-Legendre quadrature on the interval $[0, 1]$ can be used for computing the Volterra integral as follows
\begin{align*}
	v(\boldsymbol{x},t,\Psi )\approx & \sum_{j=1}^p{c_j\sum_{s=1}^{N_m}{\sum_{\iota =1}^{N_n}{w_sw_{\iota}}}\phi _{1,j}(x_1m_s)\phi _{2,j}(x_2n_{\iota})x_1x_2\left( x_1m_s-x_2n_{\iota} \right)}\\
	&\cdot \sum_{r=1}^{N_h}{w_rx_3\phi _{3,j}(x_3h_r)\sum_{k=1}^{N_{\eta}}{w_k\eta _{k}^{1+\mu}t^{2+\mu}\phi _{t,j}(t\eta _k)}},
\end{align*}
where $\left\{ w_s,m_s\right\}_{s=1}^{N_m}$, $\left\{w_\iota, m_\iota \right\}_{\iota=1}^{N_n}$, 
$\left\{ w_r, m_r\right\}_{r=1}^{N_h}$ and $\left\{ w_k,m_k \right\}_{k=1}^{N_{\eta}}$  denote the nodes and weights of the Gauss-Legendre quadrature in interval $[0,1]$, which is divided into 10 subintervals, with 16 Gauss points selected within each subinterval to ensure sufficient accuracy for discretizing the quadruple Volterra integrals.

\begin{figure}[htb!] 
	\centering
	\includegraphics[width=11.8cm,height=4.55cm]{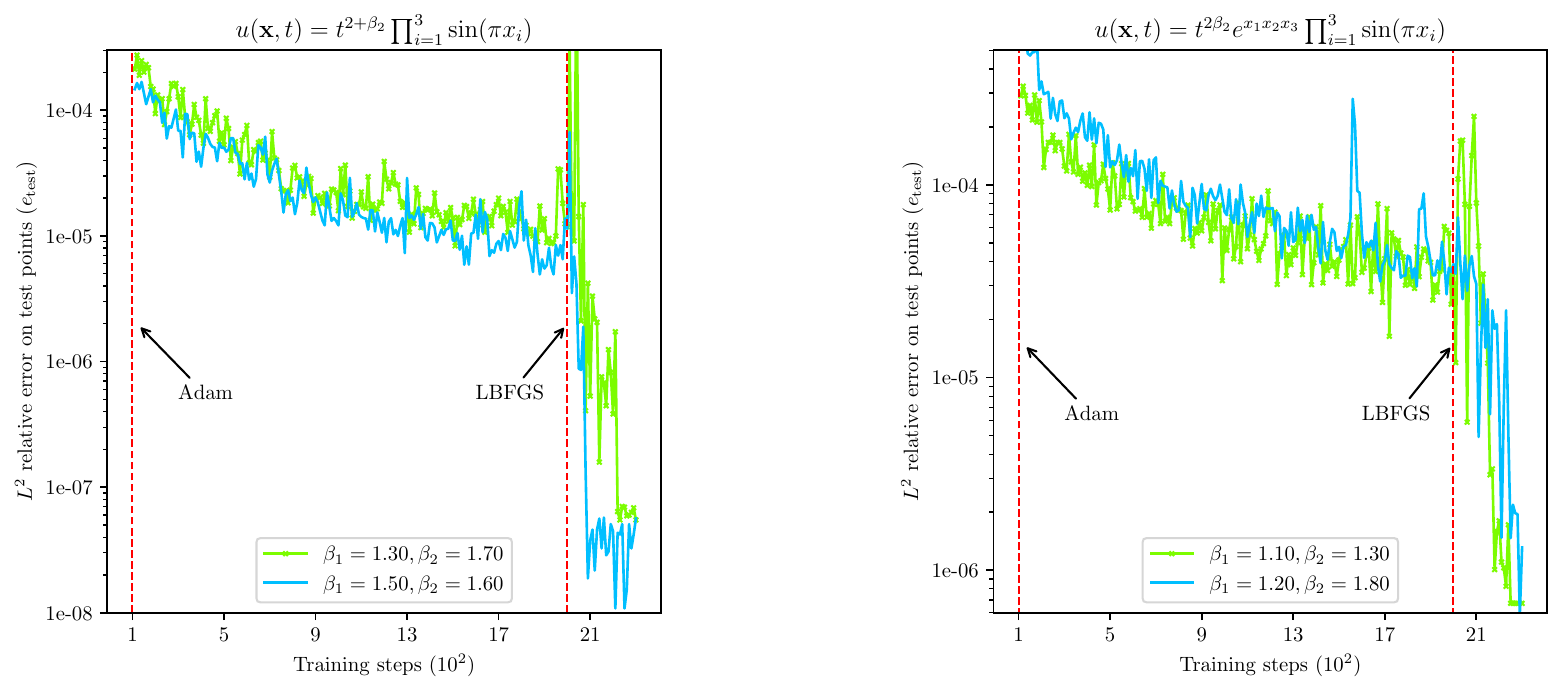}
	\caption{The loss curve and relative $L^2$ error  during the
		training process for solving \eqref{eq:3d_Volterra}.}\label{fig:ex_3d_Volterra} 
\end{figure}

\begin{table}[htb]
	\centering
	\caption{Errors of equation \eqref{eq:3d_Volterra} for different values of  $\beta_1,\beta_2 $ using TNN ($\mu = \beta_{1}$).}\label{tab:ex_3d_Volterra} 
	\begin{tabular}{cccc} 
		\toprule
		\multicolumn{2}{c}{$ u=t^{2+\beta_2}\prod_{i=1}^3{\sin(\pi x_i})$} & \multicolumn{2}{c}{$u=t^{2\beta_2}e^{x_1x_2x_3}\prod_{i=1}^3{\sin(\pi x_i})$} \\ 
		\cmidrule(lr){1-2} \cmidrule(lr){3-4}
		$(\beta_1,\beta_2)$  & $e_{\rm test}$ & $(\beta_1,\beta_2)$   & $e_{\rm test}$ \\  
		\midrule
		(1.30,1.70)     &  5.496e-08  &   (1.10,1.30)     & 6.719e-07 \\
		(1.50,1.60)     &  3.758e-08   &  (1.20,1.80)     & 8.306e-07 \\
		\bottomrule
	\end{tabular}
\end{table}

During the training process, the TNN is optimized using the Adam optimizer with an initial learning rate of 0.003 for the 2000 epochs, followed by the LBFGS optimizer with a learning rate of 0.1 for 300 steps to obtain the final result. The corresponding numerical results are shown in Figure \ref{fig:ex_3d_Volterra} and Table \ref{tab:ex_3d_Volterra}, which indicate that  the relative errors are exceptionally high, reaching up to order $10^{-7}$. 

\begin{remark}
	In this paper, we've tested the performance of TNN-based method for TFPIDEs with spatial dimension up to three. Given the successful application of the TNN-based method to other high-dimensional problems \cite{2023Tackling,wang2022tensor,wang2024solving,wang2024computing}, we believe the proposed algorithm has the potential to be extended to high-dimensional TFPIDEs (e.g., 10D or 100D) as well, which will be considered in our future work.  
\end{remark}

\section{Conclusion}
In this paper, inspired by the concept of subspace approximation from the FEM and TNN for high dimensional problems, we propose an TNN subspace-based machine learning method for solving linear and nonlinear 
TFPIDE with high accurcy and high efficiency.
Specially, we design the trial function as the TNN function multiplied by $t^{\mu}$, 
where the power $\mu$ is related to the order of the associated fractional  derivatives. The use of Gauss-Jacobi quadrature for computing the Caputo derivatives and the alternating optimization strategy both contribute to the overall accuracy.
The proposed TNN-based machine learning method demonstrates superior performance compared to fPINN and its variants, achieving significant improvements in both accuracy and computational efficiency.

In the future, we will consider the TNN-based machine learning method for 
solving other types of Fredholm or Volterra integral equations,  spatial-fractional partial differential equations as well as high-dimensional FPDEs.


\bibliographystyle{plain}
\bibliography{ref}

\end{document}